\documentclass[11pt]{article}

\usepackage[normalem]{ulem}

\usepackage[utf8]{inputenc}
\usepackage[T1]{fontenc}
\usepackage[english]{babel}
\usepackage{lmodern}

\usepackage{amsmath,amsfonts,amssymb}
\usepackage{amsthm}
\usepackage{thm-restate}
\usepackage{mathtools}
\usepackage{mathrsfs}
\usepackage{bm}
\usepackage{booktabs}
\usepackage[inline,shortlabels]{enumitem}
\usepackage[ruled,vlined,linesnumbered,algo2e]{algorithm2e}
\usepackage{tikz}
\usetikzlibrary{arrows.meta,calc,decorations.pathreplacing}

\definecolor{DarkRed}{rgb}{0.368,0.097,0.078}
\definecolor{DarkBlue}{rgb}{0,0,0.5}
\usepackage[
    colorlinks=true,
    linkcolor=DarkBlue,
    citecolor=DarkBlue,
    urlcolor=DarkRed,
    pagebackref
]{hyperref}

\usepackage[nameinlink,capitalise,noabbrev]{cleveref}
\usepackage[square,numbers]{natbib}
\usepackage{fullpage}
\usepackage{refcount}

\renewcommand{\tilde}{\widetilde}

\DeclareMathOperator*{\argmin}{arg\,min}

\DeclareMathOperator{\Reg}{Reg}
\DeclareMathOperator{\SocReg}{SocReg}
\DeclareMathOperator{\DualGap}{DualGap}

\newcommand{\N}{\mathbb{N}}

\newcommand{\R}{\mathbb{R}}

\newcommand{\cA}{\mathcal{A}}
\newcommand{\cB}{\mathcal{B}}

\newcommand{\cJ}{\mathcal{J}}

\newcommand{\cO}{\mathcal{O}}

\newcommand{\cS}{\mathcal{S}}

\newcommand{\cW}{\mathcal{W}}
\newcommand{\cX}{\mathcal{X}}
\newcommand{\cY}{\mathcal{Y}}

\newcommand{\order}{O}
\newcommand{\Ind}{\mathbb{I}}

\newcommand{\lrb}[1]{\left(#1\right)}
\newcommand{\brb}[1]{\bigl(#1\bigr)}
\newcommand{\Brb}[1]{\Bigl(#1\Bigr)}
\newcommand{\bbrb}[1]{\biggl(#1\biggr)}

\newcommand{\lcb}[1]{\left\{#1\right\}}

\newcommand{\Bcb}[1]{\Bigl\{#1\Bigr\}}
\newcommand{\bbcb}[1]{\biggl\{#1\biggr\}}

\renewcommand{\ln}{\log}

\DeclarePairedDelimiter{\inner}{\langle}{\rangle}

\DeclarePairedDelimiter{\norm}{\lVert}{\rVert}
\DeclarePairedDelimiter{\dnorm}{\lVert}{\rVert_{*}}
\DeclarePairedDelimiter{\abs}{\lvert}{\rvert}
\newcommand{\defeq}{\coloneq}
\newcommand{\eqdef}{\eqcolon}

\newcommand{\D}{D_{\psi}}

\newcommand{\dtot}{d_{\mathrm{tot}}}
\newcommand{\kappatot}{\kappa_{\mathrm{tot}}}
\newcommand{\smax}{\sigma_{\max}}
\newcommand{\qmax}{q_{\max}}

\newcommand{\MAO}{\textnormal{\textsc{MAO-OMD}}\xspace}
\newcommand{\hist}{\ensuremath{\mathcal{O}}}
\newcommand{\miss}{\ensuremath{\mathcal{P}}}
\newcommand{\obs}{\ensuremath{\mathcal{N}}}
\newcommand{\act}{\ensuremath{\mathcal{A}}}

\newcommand{\sumtT}{\sum_{t=1}^{T}}

\newcommand{\cc}[2]{c_{#1}^{(#2)}}
\newcommand{\hcc}[2]{\widehat{c}_{#1}^{(#2)}}
\newcommand{\ccdel}[2]{\alpha_{#1}^{(#2)}}
\newcommand{\ccrem}[2]{r_{#1}^{(#2)}}
\newcommand{\flow}{A}
\newcommand{\inflow}{\flow^{\mathrm{in}}}
\newcommand{\outflow}{\flow^{\mathrm{out}}}
\newcommand{\hflow}{\widehat{A}}
\newcommand{\hinflow}{\hflow^{\mathrm{in}}}
\newcommand{\houtflow}{\hflow^{\mathrm{out}}}

\SetKwInput{KwInput}{Input}
\SetKwInput{KwInit}{Initialize}
\SetKwFor{For}{for}{do}{end}
\DontPrintSemicolon

\newcommand{\tmm}{{t-1}}
\newcommand{\tpp}{{t+1}}
\newcommand{\hs}{h_{s}}
\newcommand{\gs}{g_{s}}
\newcommand{\gt}{g_{t}}
\newcommand{\w}{w}
\newcommand{\wt}{\w_{t}}
\newcommand{\wtpp}{\w_{\tpp}}
\newcommand{\cmp}{u}
\newcommand{\half}{\frac{1}{2}}
\newcommand{\brac}[1]{\left(#1\right)}
\newcommand{\sbrac}[1]{\left[#1\right]}
\newcommand{\ws}{\w_{s}}
\newcommand{\htpp}{h_{\tpp}}
\newcommand{\Inner}[1]{\left\langle#1\right\rangle}
\newcommand{\Set}[1]{\left\{#1\right\}}
\newcommand{\grad}{\nabla}

\newtheoremstyle{manualtheoremstyle}
  {3pt}      %
  {3pt}      %
  {\itshape} %
  {}         %
  {\normalfont\bfseries} %
  {.}        %
  {.5em}     %
  {}         %

\theoremstyle{manualtheoremstyle}
\newtheorem{manualtheoreminner}{Theorem}
\newenvironment{manualtheorem}[1]{%
  \renewcommand\themanualtheoreminner{#1}%
  \manualtheoreminner
}{\endmanualtheoreminner}
\crefname{manualtheoreminner}{theorem}{theorems}
\Crefname{manualtheoreminner}{Theorem}{Theorems}

\newcommand{\boldAlg}{\cB}



\allowdisplaybreaks

\theoremstyle{definition}
\newtheorem*{examplex*}{Example}

\newenvironment{example*}
  {\pushQED{\qed}\begin{examplex*}}
  {\popQED\end{examplex*}}

\theoremstyle{plain}

\newtheorem*{claim*}{Claim}
\newtheorem{lemma}{Lemma}
\newtheorem*{lemma*}{Lemma}
\newtheorem{theorem}{Theorem}
\newtheorem*{theorem*}{Theorem}

\newtheorem*{remark*}{Remark}
\newtheorem{corollary}{Corollary}
\newtheorem{fact}{Fact}
\newtheorem*{fact*}{Fact}
\newtheorem{assumption}{Assumption}
\newtheorem{definition}{Definition}
\newtheorem{proposition}{Proposition}

\newtheorem*{observation*}{Observation}

\crefname{fact}{Fact}{Facts}\Crefname{fact}{Fact}{Facts}
\crefname{assumption}{Assumption}{Assumptions}\Crefname{assumption}{Assumption}{Assumptions}
\crefname{remark}{Remark}{Remarks}\Crefname{remark}{Remark}{Remarks}

\title{Optimal Learning in Games with Arbitrary Delays:\\ Best-of-Both-Worlds for Self-Play and Adversarial Opponents}
\author{%
Emmanuel Esposito\thanks{Authors are listed in alphabetical order, equal contribution.}\\
FairPlay Team\\
CREST, ENSAE, IP Paris\\
\texttt{emmanuel@emmanuelesposito.it}
\and
Andrew Jacobsen\footnotemark[1]\\
Universitat Pompeu Fabra\\
\texttt{contact@andrew-jacobsen.com}
\and
Hao Qiu\footnotemark[1]\\
National University of Singapore\\
\texttt{qiuhaosai@gmail.com}
\and
Alexander Ryabchenko\footnotemark[1]\\
University of Toronto, Vector Institute\\
\texttt{alex.rbch.research@gmail.com}
\and
Mengxiao Zhang\footnotemark[1]\\
University of Iowa\\
\texttt{mengxiao-zhang@uiowa.edu}
}
\date{}

\begin{document}

\hypersetup{pageanchor=false}
\pagenumbering{roman}
\maketitle
\thispagestyle{empty}

\begin{abstract}
  We study learning in zero-sum games with delayed feedback. We introduce \emph{Multi-Anchor Optimistic Online Mirror Descent} (MAO-OMD), a single algorithm which obtains optimal fast rates in self-play on smooth games, while simultaneously maintaining minimax optimal guarantees against arbitrary opponents. In particular, in smooth zero-sum games with arbitrary and unknown feedback delays, MAO-OMD obtains $O(\sigma_{\max}+1)$ social and individual regrets when used by both players, where $\sigma_{\max}$ denotes the maximum number of outstanding gradients, while also guaranteeing minimax optimal $O(\sqrt{T+d_{\mathrm{tot}}})$ individual regret against arbitrary opponents, where $d_{\mathrm{tot}}$ is the total delay. Moreover, we show that these optimal rates cannot be obtained simultaneously using the standard black-box reduction to the undelayed setting. 
  We also extend our results to multi-player games, and to games with player-dependent delay sequences.
\end{abstract}

\clearpage
\tableofcontents

\clearpage
\pagenumbering{arabic}
\hypersetup{pageanchor=true}

\section{Introduction}
\label{sec:intro}

In this paper, we study learning in two-player zero-sum games with delayed feedback over $T$ rounds.
Formally, let $f:\cX\times\cY\to\R$ be a smooth convex--concave function on nonempty, compact, convex action sets $\cX$ and $\cY$.
In each round $t$, the players simultaneously choose $x_t\in\cX$ and $y_t\in\cY$, and face the convex loss functions $x\mapsto f(x,y_t)$ and $y\mapsto-f(x_t,y)$, respectively.
The players observe their respective loss gradients $\nabla_x f(x_t,y_t)$ and $-\nabla_y f(x_t,y_t)$ at the end of round $t+d_t$, where the \emph{delay} $d_t$ is unknown in advance and may vary across rounds. 
We measure each player's performance by its \emph{individual (external) regret}, the difference between its cumulative loss and that of the best fixed action chosen in hindsight against the opponent's realized actions. We also consider \emph{social regret}, defined as the sum of the players' individual regrets. 

In the absence of delays, individual and social regret have been studied extensively \citep{fudenberg1998theory, cesa2006prediction, nisan2007algorithmic, shalev2012online}.
A classical result linking regret minimization to equilibrium computation shows that, in a two-player zero-sum game, a social regret bound of $R_T$ guarantees that the players' averaged strategies form an $(R_T/T)$-approximate Nash equilibrium \citep{freund1999adaptive}.
Thus, standard no-regret algorithms with $\order(\sqrt T)$ regret automatically yield an $\order(1/\sqrt T)$ convergence rate. 
However, it is known that significantly faster rates can be achieved in \emph{self-play}, where both players
employ the same learning algorithm. In particular,
when both players employ \emph{optimistic} algorithms, which use a prediction of the forthcoming gradient when choosing the next action,
the resulting dynamics can induce significantly faster convergence to equilibrium \citep{chiang2012,rakhlin2013predictable,rakhlin2013games,syrgkanis2015fast}.
In zero-sum matrix games, \cite{rakhlin2013games} established an $\order((\log T)/T)$ equilibrium convergence rate for optimistic mirror descent, while \cite{syrgkanis2015fast} further showed that optimistic algorithms achieve $\order(1)$ social regret in general-sum games, yielding an $\order(1/T)$ convergence rate in the zero-sum case. In general-sum games, \cite{chen2020hedging} obtained improved bounds on external and swap regret, while \cite{daskalakis2021nearoptimal,anagnostides2022} proved bounds that depend only polylogarithmically on $T$. Recent preprints have further established individual or swap regret bounds that are independent of $T$ in finite general-sum games \citep{liu2026constant,abbadi2026constant,soleymani2026multiplicative,mai2026constant}.

However, the understanding of fast regret rates under delayed feedback remains limited. The work of \cite{fujimoto2025} showed that even a fixed one-round delay, $d_t=1$ for all $t$, can destroy the constant social-regret guarantee of optimistic Follow-the-Regularized-Leader (FTRL).
For fixed delays $d_t=d$ for all $t$, they recovered $\order((d+1)^2)$ social regret by scaling the weight of the optimistic term in the update by a factor of $d+1$. More recently, \cite{zhuang2026optimal} obtained $\tilde O(d+1)$ social and individual regret in two-player zero-sum games by interleaving $d+1$ independent copies of optimistic FTRL, and proved a matching $\Omega(d+1)$ lower bound. Both approaches are formulated for a fixed, known delay $d$, with the latter requiring both players to follow the same interleaving schedule determined by $d$. These results leave open how to obtain sharp regret guarantees in self-play that adapt to round-dependent delays unknown in advance, while retaining optimal regret against any opponent. We therefore ask:

\begin{center}
\emph{Can a learning algorithm achieve the optimal social regret in self-play under arbitrary time-varying delays, while simultaneously guaranteeing the optimal regret against an arbitrary opponent?}
\end{center}

\subsection{Our results}

We answer this question affirmatively. The following informal theorem summarizes our main results, which are stated precisely in \Cref{sec:main-results} (\Cref{thm:main-games,thm:main-robust}).

\begin{quote}
\vskip-2em
\begin{theorem*}[informal]
For smooth convex--concave zero-sum games with arbitrary unknown delays, there exists a learning algorithm that guarantees every player using it $\order(\sqrt{T+\dtot})$ regret against an arbitrary opponent, improving to $\order(\smax+1)$ when the opponent uses the same algorithm.
Here $\smax$ is the maximum number of earlier rounds whose feedback is missing at any point in time, and $\dtot=\sum_{t=1}^Td_t$ is the total delay.
\end{theorem*}
\end{quote}
Both of these rates are optimal.
Under fixed delays $d_t\equiv d$, we have $\smax=d$, and \cite[Theorem~8]{zhuang2026optimal} shows that for every learning dynamic there is a smooth zero-sum game in which each player suffers $\Omega(d+1)$ regret in self-play, so the dependence on $\smax$ cannot be improved.
Against an arbitrary opponent, a player faces online convex optimization with delayed feedback, for which $\sqrt{T+\dtot}$ is the minimax regret rate established by~\cite{quanrud2015}.

To achieve these guarantees, we introduce \emph{Multi-Anchor Optimistic Online Mirror Descent} (\MAO), an optimistic mirror descent method designed to adapt to the delays as they unfold, without any prior knowledge about them.
The key challenge in extending existing optimistic methods to delayed feedback is that delays break the mechanism behind their fast rates.
Without delays, the prediction error of each round is controlled by the movement between consecutive actions; the stability term of the update, which regularizes toward the previous action, pays for exactly that movement. Under delays, however, a newly observed gradient may have been generated at an action played many rounds earlier, so its prediction error is controlled by the movement from that past action to the current one, which the stability term toward the previous action does not cover.
To account for this, \MAO regularizes each update toward multiple past actions, its \emph{anchors}, including the actions whose gradients have just been observed. Such regularization produces the stability terms that pay for the prediction errors of the newly observed gradients, while retaining the negative movement terms needed for fast convergence in games.

Choosing these weights is the main technical challenge in the design of \MAO, because they must do two things at once.
On the one hand, the weight on each anchor must be large enough to pay for the corresponding prediction error.
On the other hand, with arbitrary delays feedback arrives out of order and an action is anchored at several rounds, which destroys the telescoping structure of standard mirror-descent regret bounds. The cancellation survives only if no action is anchored with more total weight than it was computed with.
Inspired by the weighting principle of Banker-OMD \citep{pmlr-v202-huang23e,qiu2026universal}, we develop an adaptive allocation of regularization weights across pending rounds that meets both conditions and yields a regret bound whose prediction-error costs scale with the total delay while keeping the negative stability terms needed for fast rates.

The paper is organized as follows.
\Cref{sec:prelim} introduces the setting and the notation, and \Cref{sec:main-results} states our main results.
In \Cref{sec:bold}, we provide a variant of the BOLD reduction~\cite{joulani2013} achieving $\order(\smax+1)$ regret in games with arbitrary delays, and show that no such variant can bring its regret against an arbitrary opponent within a factor $\sqrt{1+\ln(\smax+1)}$ of the optimal rate, which motivates \MAO.
\Cref{sec:mao_omd,sec:analysis} introduce \MAO together with its weight-allocation strategy, and contain its analysis, respectively.
Finally, \Cref{sec:multi,sec:hetero} extend our main results to multi-player games and to player-dependent delays, respectively. Additional related work on learning with delayed feedback and in games is discussed in \Cref{app:related-work}.

\section{Setting and notation}
\label{sec:prelim}

Let $\N \defeq \{0,1,2,\dots\}$ denote the natural numbers and let $\N_+ \defeq \N \setminus \{0\}$.
For an integer $n\in\N$ let $[n]\defeq\{1,\dots,n\}$, with $[0]=\emptyset$.
The disjoint union of sets is denoted by $\sqcup$.

We first introduce a general online protocol with delayed feedback, which describes the interaction from an individual player's perspective. We then specialize this protocol to two-player zero-sum games. Finally, we introduce the delay-induced notation used throughout the analysis.

\paragraph{Online protocol.}
Let $\cW\subseteq\R^n$, $n \in \N_+$ be a nonempty, compact, and convex domain, equipped with a norm $\norm{\cdot}$ and its dual norm $\dnorm{\cdot}$. The interaction proceeds over $T\in\N_+$ rounds.
At the start of each round $t\in[T]$, the player selects an action $w_t\in\cW$ and the environment then chooses a differentiable convex loss function $f_t:\cW\to\R$, resulting in a loss $f_t(w_t)$ that the player does not observe.
Feedback for round $t$ arrives with a delay: at the end of each round $t'\ge t$, the environment decides whether to release the feedback for round $t$, and if it does, it reveals a gradient $g_t :=\grad f_t(w_t)$. The feedback for every round is released exactly once, and we denote the resulting delay by $d_t\in\N$, so that $g_t$ is revealed at the end of round $t+d_t\le T$.\footnote{Feedback arriving after round $T$ does not affect the learner's actions, so we may assume $d_t\le T-t$ for all $t\in[T]$.}

Throughout, every environment's choice may depend on the entire history up to that point.
Accordingly, our bounds depend on realized loss functions $f_t$, gradients $g_t$, and delays $d_t$.
The performance is measured by the external regret with respect to a comparator $u\in\cW$:
\begin{equation*}
    \Reg_T(u) \defeq \sumtT \brb{f_t(w_t) - f_t(u)} \;,
    \qquad
    \Reg_T \defeq \sup_{u \in \cW} \Reg_T(u) \;, 
\end{equation*}
where $\Reg_T$ denotes the regret against the best fixed comparator in hindsight. By convexity, the external regret above is bounded by the \emph{linearized regret}: $\Reg_T(u) \le  \sumtT \inner{g_t, w_t-u}$.

\paragraph{Zero-sum games.}
In two-player zero-sum games, each player faces the online protocol above with loss functions determined by the other player's decisions. 
Let $\cX\subseteq\R^{n_x}$ and $\cY\subseteq\R^{n_y}$ be nonempty, compact, and convex domains, and let $f:\cX\times\cY\to\R$ be differentiable, convex in its first argument, and concave in its second. At round $t\in[T]$, the players simultaneously select $x_t\in\cX$ and $y_t\in\cY$, incurring losses $f(x_t,y_t)$ and $-f(x_t,y_t)$. The players receive their loss gradients $g_t^x \defeq \nabla_x f(x_t,y_t)$ and $g_t^y \defeq -\nabla_y f(x_t,y_t)$ at the end of round $t+d_t$.
Their \emph{individual regrets} are
\begin{align*}
    \Reg_T^x \defeq \sup_{x\in\cX} \sumtT
    \brb{f(x_t,y_t)-f(x,y_t)} \;,\qquad
    \Reg_T^y \defeq \sup_{y\in\cY} \sumtT
    \brb{f(x_t,y)-f(x_t,y_t)} \;,
\end{align*}
and their \emph{social regret} is
$\SocReg_T \defeq \Reg_T^x+\Reg_T^y$.
By convexity, the social regret can be used to control the duality gap of the averaged decisions
$\bar x_T\defeq \frac{1}{T}\sumtT x_t$ and
$\bar y_T\defeq \frac{1}{T}\sumtT y_t$:
\begin{equation}
   \DualGap_T \defeq \sup_{y\in\cY} f(\bar x_T,y) - \inf_{x\in\cX} f(x,\bar y_T)
   \le
   \frac{1}{T}\SocReg_T \;,\label{eq:gap-to-social-regret}
\end{equation}
which measures the deviation from the saddle point of $f$, i.e., a \emph{Nash equilibrium} \citep{freund1999adaptive,farina2019online}.

\paragraph{Delay-induced notation.}
Throughout the paper, we use the following notation induced by the sequence of realized delays $(d_t)_{t=1}^T$. For each round $t\in[T]$, define four sets of round indices:
\begin{align*}
  \text{(observed) }\,\hist_t &\defeq \{s\in[t-1]:s+d_s<t\}, &
  \text{(pending) }\,\miss_t &\defeq \{s\in[t-1]:s+d_s\ge t\},\\
  \text{(new) }\,\,\quad\obs_t &\defeq \{s\in[t]:s+d_s=t\}, &
  \text{(active) }\,\,\,\act_t &\defeq \miss_t\cup\{t\}.
\end{align*}
The sets $\hist_t$ and $\miss_t$ partition the past rounds into those whose feedback has been \emph{observed} before round $t$ and those whose feedback remains \emph{pending} at its start. The set $\obs_t$ contains rounds whose feedback is \emph{newly observed} in round $t$, while the \emph{active} set $\act_t$ consists of all rounds $s\in[t]$ whose feedback arrives in round $t$ or later. In particular, $\hist_t \sqcup \miss_t = [t-1]$, $\bigsqcup_{t=1}^T \obs_t = [T]$, and $\obs_t\subseteq\act_t$.

We write $q_t\defeq\abs{\act_t}$ and $\sigma_t\defeq\abs{\miss_t}$ for the numbers of active and pending rounds, respectively, and denote their maxima by $\qmax\defeq\max_{t\in[T]}q_t$ and $\smax\defeq\max_{t\in[T]}\sigma_t$, the \emph{maximum numbers of active and pending rounds}, and the \emph{total delay} by $\dtot\defeq\sum_{t\in[T]}d_t$.
Since $q_t=\sigma_t+1$, we have $\qmax=\smax+1$.
Moreover, as each round $t$ is pending during exactly $d_t$ rounds, $\sum_{t\in[T]}\sigma_t=\dtot$ and hence $\sum_{t\in[T]}q_t=T+\dtot$.
The maximum number of active rounds is also controlled by the total delay: $\qmax\le\sqrt{T+2\dtot}$ (\Cref{fact:sets}(e)).
As is standard in the delayed-feedback literature, our bounds are stated in terms of $\smax$ and $\dtot$, with $\qmax$ used in place of $\smax+1$ for convenience.

We also use the notion of \emph{congestion}, defined as the maximum number of active rounds while a given round is active.
Concretely, for $t\in[T]$, $\kappa_t\defeq \max_{t\le r\le t+d_t}q_r$ and $\kappa_{t,s}\defeq\max_{t\le r\le s}q_r$ for $t\le s\le t+d_t$, where $\kappa_t$ is the congestion of round $t$ and $\kappa_{t,s}$ is its running version, observable at round $s \le t + d_t$.
An important property of congestion is that its total sum $\kappatot \defeq \sum_{t\in[T]}\kappa_t$ is of order $\Theta(T + \dtot)$. Indeed, any round active between $t$ and $t+d_t$ is either already active at round $t$ or is one of the rounds $t+1,\dots,t+d_t$, so that $q_t\le\kappa_t\le q_t+d_t$ and hence $T+\dtot\le\kappatot\le T+2\dtot$.
We use this property in the analysis: by charging each round roughly in proportion to its own congestion rather than the global $\qmax$, we obtain bounds that scale with $T+\dtot$ instead of $T\qmax$.

Further useful properties of these delay-induced quantities are collected in \Cref{app:delay}.

\section{Main results}
\label{sec:main-results}

In this section, we formally state the guarantees of our algorithm, \MAO, in smooth convex--concave zero-sum games: fast rates when both players run this algorithm, and the optimal regret under delays for a player that runs it against an arbitrary opponent.

For the two-player zero-sum game of \Cref{sec:prelim}, let $g^x(x,y)\defeq\nabla_xf(x,y)$ and $g^y(x,y)\defeq-\nabla_yf(x,y)$ denote the players' loss gradients at a profile $(x,y)\in\cX\times\cY$, so that $g^p_t=g^p(x_t,y_t)$ for $p\in\{x,y\}$.
Following~\cite{rakhlin2013games}, we assume that $f$ is Lipschitz and smooth in the profile, that is, both gradient maps are bounded and Lipschitz, with constants known to the players (\Cref{ass:regularity}). 

\begin{assumption}[$G$-Lipschitzness and $L$-smoothness]
\label{ass:regularity}\label{ass:smooth}
There exist constants $G, L > 0$ such that
\begin{equation*}
\dnorm{g^p(x,y)}\leq G
\qquad\text{and}\qquad
\dnorm{g^p(x,y)-g^p(x',y')} \le L\brb{\norm{x-x'}+\norm{y-y'}}
\end{equation*}
for each player $p\in\{x,y\}$ and all $x,x'\in\cX$ and $y,y'\in\cY$. 
\end{assumption}

\MAO is an optimistic online mirror descent (OMD) method,
which we describe for a player facing the online protocol of \Cref{sec:prelim}, with action set $\cW$, actions $w_t$, and loss gradients $g_t$.
The method takes as input a regularizer $\psi:\cW\to\R$ and a reference point $w_0\in\cW$, and produces the iterates $w_1,w_2,\dots$ starting from $w_0$: each iterate $w_{t+1}$ is computed by a mirror descent step that, as in every OMD method, is regularized by the Bregman divergence $D_\psi(u\mid w)\defeq\psi(u)-\psi(w)-\inner{\nabla\psi(w),u-w}$, here taken with respect to several of the previous iterates $w_0,\dots,w_t$.
As in every optimistic method \cite{chiang2012,rakhlin2013predictable}, each iterate $w_t$ is selected using a hint $h_t$ for the upcoming gradient $g_t$, formed from the feedback revealed before round $t$.
Although \MAO{} regularizes toward many previous iterates, our analysis only requires the Bregman radius of $\cW$ around $w_0$ to be bounded (\Cref{ass:diam}).

\begin{assumption}[Strongly convex regularizer and bounded Bregman radius] \label{ass:diam}
The regularizer $\psi$ is differentiable and $1$-strongly convex with respect to $\norm{\cdot}$, i.e., $D_\psi(u\mid w)\ge\frac12\norm{u-w}^2$ for all $u,w\in\cW$.
Additionally, there is a known constant $M>0$ and reference point $w_0\in\cW$ such that $D_\psi(w\mid w_0)\le M$ for all $w\in\cW$. 
\end{assumption}

At each round $t$, \MAO{} uses as its hint the earliest observed gradient that has not yet served as a hint.
Specifically, let $s_t$ be the smallest element of $\hist_t\setminus\{s_1,\ldots,s_{t-1}\}$, with $s_t=0$ if this set is empty, and set $h_t\defeq g_{s_t}$, where $g_0\defeq0$.
By construction, each observed gradient serves as a hint at most once, and at most $\qmax=\smax+1$ rounds have $s_t=0$ (as shown in \Cref{lem:queue}).
The following theorem states the main guarantee for a player running \MAO.

\begin{restatable}[informal]{theorem}{MainOptimistic}\label{thm:main-optimistic}
Let Assumptions~\ref{ass:regularity} and \ref{ass:diam} hold.
For arbitrary unknown delays and any sequence of opponent's actions, a player that runs \MAO (\Cref{alg:mao}) 
guarantees that, for all $u\in\cW$,
\begin{equation*}
    \sumtT\inner{g_t,w_t-u}
    \le
    \brb{8LM + 4G\sqrt{M}}\qmax + 2\sqrt{2M \sumtT \kappa_t\dnorm{g_t - h_t}^2} - L\sumtT \norm{w_t - w_{s_t}}^2 \;.
\end{equation*}
\end{restatable}
We state the complete version of \Cref{thm:main-optimistic} in \Cref{sec:algorithm} and provide its proof in \Cref{sec:main-optimistic}.
In the remainder of this section, we derive its implications for two-player zero-sum games, which are our main results.

\paragraph{Fast rates when both players run MAO-OMD.}
Suppose that both players run \MAO with regularizers $\psi_x$ and $\psi_y$ satisfying \Cref{ass:diam} with the same constant $M$.
Since the hint rule depends only on the delays, which are common to both players, they select the same hint source $s_t$ at every round, and whenever $s_t>0$ their hints $h^p_t=g^p(x_{s_t},y_{s_t})$ come from the same past profile.
Smoothness therefore bounds each player's prediction error by the movement of both players since round $s_t$, which the negative terms of \Cref{thm:main-optimistic}, summed over the two players, control.
\Cref{thm:main-games} uses this to bound the social regret, and the nonnegativity of the social regret in zero-sum games extends the bound to individual regrets.

\begin{theorem}\label{thm:main-games}
    Let \Cref{ass:regularity} hold.
    If both players run \MAO with regularizers 
    satisfying \Cref{ass:diam} under arbitrary unknown delays, then the social regret $\SocReg_T$ and individual regrets $\Reg_T^x$, $\Reg_T^y$ are all $\order\bigl((G\sqrt M+LM)\qmax\bigr)$, while $\DualGap_T$ is $O\Bigl(\frac{(G\sqrt{M}+LM)q_{\max}}{T}\Bigr)$.
\end{theorem}
\begin{proof}
    Let $\cW_x=\cX$ and $\cW_y=\cY$. Denote by $\Delta^p_T\defeq\sum_{t=1}^T\norm{w^p_t-w^p_{s_t}}^2$ the movement of player $p\in\{x,y\}$ relative to the hint sources, where $w^x_t=x_t$ and $w^y_t=y_t$, and let $\Delta_T\defeq\Delta^x_T+\Delta^y_T$.
    Both players use the same hint sources $s_t$, so at every round $t$ with $s_t>0$, $L$-smoothness gives $\dnorm{g^p_t-h^p_t}^2\le L^2\brb{\norm{x_t-x_{s_t}}+\norm{y_t-y_{s_t}}}^2$.
    At every round $t$ with $s_t=0$, $h^p_t=0$ and $G$-Lipschitzness gives $\dnorm{g^p_t-h^p_t}^2\le G^2$ instead. As there are at most $\qmax$ rounds with $s_t = 0$ (see \Cref{lem:queue} provided later in \Cref{sec:analysis} plus the choice $s_1=0$) and $\max_t\kappa_t\le\qmax$, we have $\sumtT\kappa_t\dnorm{g_t^p-h_t^p}^2 \le \qmax\brb{G^2\qmax+2L^2\Delta_T}$.
    \Cref{thm:main-optimistic} then gives, for all $u\in\cW_p$,
    \begin{align*}
        \sumtT\inner{g^p_t,w^p_t-u}
        &\le
        \brb{8LM+4G\sqrt M}\qmax+2\sqrt{2M\qmax\brb{G^2\qmax+2L^2\Delta_T}}-L\Delta^p_T\\
        &\le
        \brb{8LM+(4+4)G\sqrt M}\qmax+4L\sqrt{M\qmax\Delta_T}-L\Delta^p_T \;,
    \end{align*}
    where the last inequality uses $\sqrt{a+b}\le\sqrt a+\sqrt b$.
    By convexity of $f(\cdot,y_t)$ and concavity of $f(x_t,\cdot)$, the left-hand side upper bounds $\Reg_T^p(u)$.
    Taking the supremum over $u$ bounds the individual regrets, and summing the resulting bounds for the two players bounds the social regret:
    \begin{align}
        \Reg_T^p
        &\le
        8\brb{LM+G\sqrt M}\qmax+4L\sqrt{M\qmax\Delta_T}-L\Delta^p_T \;,\label{eq:main-player}\\
        \SocReg_T
        &\le
        16\brb{LM+G\sqrt M}\qmax+8L\sqrt{M\qmax\Delta_T}-L\Delta_T \;.\label{eq:main-social}
    \end{align}
    By the AM-GM inequality, we further have that $\SocReg_T\le\brb{16G\sqrt M+32LM}\qmax$, and consequently $\DualGap_T\le\SocReg_T/T \le \brb{16G\sqrt M+32LM}\,\frac{\qmax}{T}$ via \Cref{eq:gap-to-social-regret}. %

    To bound the individual regrets, we first bound $\Delta_T$.
    Since $\SocReg_T\ge T\cdot\DualGap_T\ge0$, applying $8L\sqrt{M\qmax\Delta_T}\le L\Delta_T/2+32LM\qmax$ in \eqref{eq:main-social} and rearranging gives $L\Delta_T\le\brb{32G\sqrt M+96LM}\qmax$.
    Substituting this into \eqref{eq:main-player} after applying $4L\sqrt{M\qmax\Delta_T}\le L\Delta_T/4+16LM\qmax$, and dropping the nonpositive term $-L\Delta^p_T$, we obtain $\Reg^p_T\le\brb{16G\sqrt M+48LM}\qmax$.
\end{proof}

Suppressing the dependence on $G$, $L$, and $M$, both individual and social regrets are $\order({\qmax}) = \order(\smax+1)$ and the equilibrium convergence rate is $\order({(\smax+1)}/{T})$, with the dependence on $\smax$ matching the lower bound of~\cite[Theorem~8]{zhuang2026optimal}. In the absence of delays, i.e., $d_t=0$ for all $t$, our regret bound becomes $\order(G\sqrt M+LM)$, recovering the regret of optimistic mirror descent with an optimally tuned fixed step size~\cite[Lemma~1]{rakhlin2013games}.

\paragraph{Regret against an arbitrary opponent.}
Suppose now that player $x$ runs \MAO while player $y$ takes arbitrary, potentially adversarial actions.
The hints of player $x$ need not be related to the gradients it receives, so the negative term of \Cref{thm:main-optimistic} cannot be relied on, but the prediction errors remain bounded by $G$-Lipschitzness.
\Cref{thm:main-robust} shows that this is enough for optimal regret.

\begin{theorem} \label{thm:main-robust}
    Let \Cref{ass:regularity} hold.
    If player $x$ runs \textnormal{\MAO} with a regularizer satisfying \Cref{ass:diam} under arbitrary unknown delays and against an opponent $y$ taking arbitrary actions, then its individual regret $\Reg^x_T$ is $\order\brb{G\sqrt{M(T+\dtot)}+LM\qmax}$.
\end{theorem}
\begin{proof}
    Each hint of player $x$ is either a gradient or zero, so $G$-Lipschitzness gives $\dnorm{g^x_t-h^x_t}^2\le4G^2$ for all $t$.
    Since $\sumtT\kappa_t\le T+2\dtot$ (\Cref{fact:kappa}(b)), $\sumtT\kappa_t\dnorm{g^x_t-h^x_t}^2\le8G^2(T+\dtot)$.
    \Cref{thm:main-optimistic} then gives $\Reg^x_T\le\sup_{u\in\cX}\sumtT\inner{g^x_t,x_t-u}\le\brb{8LM+4G\sqrt M}\qmax+8G\sqrt{M(T+2\dtot)}$, by convexity of $f(\cdot,y_t)$. To obtain the stated bound, use $\qmax^2 \le {T+2\dtot}$ (\Cref{fact:sets}(e)).
\end{proof}

Suppressing the dependence on $G$, $L$, and $M$, and since $\qmax^2\le T+2\dtot$ (\Cref{fact:sets}(e)), the regret bound of \Cref{thm:main-robust} is $\order(\sqrt{T+\dtot})$, which matches the optimal rate for delayed online convex optimization \citep{quanrud2015}.
Together, \Cref{thm:main-games,thm:main-robust} show that a player running \MAO obtains both guarantees in the same run, without knowing whether its opponent also runs \MAO.

\paragraph{Additional results.}
We extend \MAO to three further settings. 
For brevity, the bounds in this paragraph suppress the dependence on $G$, $L$, and $M$.
First, in \Cref{app:multi}, we consider $N$-player smooth games with nonnegative social regret, as in~\cite{anagnostides2022lastiterate}, which include zero-sum and constant-sum polymatrix games.
We show that \MAO, with an optimism weight that scales with $N$, guarantees social regret $\order\brb{N^2\qmax}$ and individual regret $\order\brb{N\qmax}$ when all players run it, and individual regret $\order\brb{N\qmax + \sqrt{T+\dtot}}$ against arbitrary opponents.
Second, in \Cref{app:hetero}, we consider player-dependent delays, where the feedback of a round may reach the two players at different times.
If the players communicate which rounds they have received, they can run \MAO under the common effective delays $\max\{d^x_t,d^y_t\}$, and we show that both social and individual regret are then $\order\brb{\qmax^x+\qmax^y}$.
Finally, in \Cref{app:oco}, we consider single-player online convex optimization, where the losses are Lipschitz and not necessarily smooth, and show that \MAO still achieves the optimal $\order\brb{\sqrt{T+\dtot}}$ regret of~\cite{quanrud2015}.

\section{A scheduling-based approach and its limitation}
\label{sec:bold}

A natural approach to learning with delayed feedback is to run multiple
independent instances of an undelayed learning algorithm, ensuring that each
instance receives feedback from its previous action before being used again.
This is the idea behind the BOLD reduction of \cite{joulani2013}, which assigns each
round $t\in[T]$ to an available instance, creating a new one whenever all
existing instances are awaiting feedback.
The resulting assignment partitions $[T]$ into subsequences $I_1,\ldots,I_{\qmax}$, such that consecutive rounds $s<t$ in the same subsequence satisfy $s+d_s<t$. Thus, each subsequence can be handled by an independent undelayed online-learner $\boldAlg$, and the total number of instances created is $\qmax$.
If the base learner
guarantees regret at most $C_{\boldAlg}\sqrt n$ on any $n$ rounds,
then, writing $n_k=|I_k|$, the standard analysis gives
$\Reg_T\le C_{\boldAlg}\sum_{k=1}^{\qmax}\sqrt{n_k}
\le C_{\boldAlg}\sqrt{\qmax T}$, 
which can be significantly larger than the minimax
$O(\sqrt{T+\dtot})$ rate \citep{quanrud2015}.

BOLD allows each round to be assigned to any available instance, i.e., one whose
previous feedback has already arrived. We refine this assignment rule by
indexing instances in order of creation and always selecting the available
instance with the smallest index, creating a new one only when none is
available. By preferentially reusing earlier instances, this greedy rule yields
a sharper regret bound, losing only a logarithmic factor relative to the minimax
rate, while still retaining the $\order(\qmax)$ rates when used by both players
of a smooth game.
The resulting guarantees are stated in the following theorem, with the pseudocode and proofs deferred to \Cref{app:assignment}.

\begin{samepage}
\begin{theorem}\label{thm:assign}
The refined BOLD reduction in \cref{alg:refined-bold} has the following guarantees without advance knowledge of the delays:
\begin{enumerate}[label=(\roman*),itemsep=0pt,leftmargin=*]
  \item In the online convex setting of \cref{sec:prelim}, if $\boldAlg$ guarantees regret at most $C_{\boldAlg}\sqrt n$ for every $n\ge1$, then
  \(
    \Reg_T\le C_{\boldAlg}\sum_k\sqrt{n_k}
    \le C_{\boldAlg}\sqrt{(T+\dtot)(1+\ln\qmax)} \,.
  \)
  \item Under \Cref{ass:smooth,ass:diam}, if $\boldAlg$ guarantees individual and social regret $O(LM+G\sqrt M)$ in undelayed zero-sum self-play for every horizon, then both players using the reduction with common delays satisfy
  \(
    \SocReg_T, \Reg_T^p
    =O\brb{(LM+G\sqrt M)\qmax} \text{ for } p\in\{x,y\}.
  \)
\end{enumerate}
\end{theorem}
\end{samepage}

The result for part~(i) follows by always choosing the smallest available index over idle instances of $\boldAlg$, which ensures  $\sum_k k n_k\le T+\dtot$,
and the weighted Cauchy-Schwarz inequality yields the stated regret bound.
For part~(ii), under common delays both players assign each round to the same instance, creating the same partition $I_1, \dots, I_{\qmax}$ of rounds, and each instance receives feedback from its previous round before being used again.
Thus, each pair of corresponding instances effectively plays the same undelayed game, and the RVU property \citep{syrgkanis2015fast} yields the desired $\order(\qmax)$ regret guarantee, as recently shown by \cite{zhuang2026optimal} for fixed delays with $\qmax = d+1$.

Although part~(ii) of \Cref{thm:assign} matches the desired $\order(\qmax)$ fast regret rate when both players run this refined version of BOLD, the additional $\sqrt{1+\ln\qmax}$ overhead against adversarial opponents in part~(i) of \Cref{thm:assign} cannot be removed by any partitioning of rounds, as we show in the following proposition.
\begin{proposition}\label{prop:assign-lower}
For every integer $\smax\ge1$, there are arbitrarily long delay sequences whose maximum number of pending rounds is exactly $\smax$ and for which every valid assignment of rounds to instances, with loads $n_k$, satisfies
\(
  \sum_k\sqrt{n_k}\ge\sqrt{(T+\dtot)\frac{1+\ln\qmax}{2}} .
\)
Moreover, for every fixed $\gamma>0$, if each instance runs the optimistic OMD learner specified in \Cref{app:assignment-omd-lower}, an adaptive linear-loss adversary forces every valid assignment to incur
\(
  \Reg_T=\Omega\Brb{\sqrt{(T+\dtot)(1+\ln\qmax)}} .
\)
The implicit constant may depend on $\gamma$ but is independent of $T$ and $\smax$.
\end{proposition}

In the next section, we develop a \emph{single-instance} algorithm, \MAO, which retains the fast $\order(\qmax)$ rate while guaranteeing $O(G\sqrt{M(T+\dtot)}+LM\qmax)$ regret against arbitrary opponents in the same run (\Cref{thm:main-games,thm:main-robust}). Since $\qmax^2\le T+2\dtot$ (via \Cref{fact:sets}), 
this gives regret $O(\sqrt{T+\dtot})$, without the additional
$\sqrt{1+\ln\qmax}$ factor in \Cref{thm:assign} part~(i).

\section{Multi-anchor optimistic online mirror descent}
\label{sec:mao_omd}

The scheduling-based approach of \Cref{sec:bold} can achieve near-optimal regret in
smooth games when applied with OMD as the base learner, 
but will incur a suboptimal logarithmic factor in its
adversarial regret guarantee as shown above in \Cref{prop:assign-lower}. To obtain both optimal guarantees
simultaneously, we develop a new optimistic learning algorithm that
handles delayed feedback directly, without maintaining independent
learning instances. The key challenge is to obtain an optimistic regret
bound that controls the prediction errors $\norm{\gt-h_{t}}^{2}_{*}$
individually, while retaining negative terms that can
compensate for these errors when both players use the algorithm in a smooth game.
Existing optimistic regret bounds for delayed feedback
do not directly provide this combination of properties, even for
fixed delays, and extending them to arbitrary delays introduces
further difficulties because gradients may arrive out of order 
\citep{flaspohler2021,fujimoto2025,zhuang2026optimal}.
In this section we introduce \emph{Multi-Anchor
Optimistic Online Mirror Descent} (\MAO), a generalization of
optimistic mirror descent framework, to address this challenge.
Our starting point is the following regret decomposition,
which expresses the regret in terms of the feedback received at each
round.
\begin{restatable}[Delays-to-feedback decomposition]{lemma}{regretdecomp} \label{lem:regret-decomp}
    For any $u \in \cW$ and any sequence $h_1, \dots, h_{T+1} \in \R^n$ of hints with $h_1 = h_{T+1} = 0$,
    \begin{equation}
        \sumtT \inner{g_t, w_t - u}
        =
        \sumtT \inner*{h_{t+1} + \sum_{s \in \obs_t} (g_s - h_s), w_{t+1} - u}
        + \sumtT \sum_{s \in \obs_t} \inner*{g_s - h_s, w_s - w_{t+1}} \;.\label{eq:feedback-decomp}
    \end{equation}
\end{restatable}

The decomposition in \Cref{lem:regret-decomp} naturally suggests an
optimistic mirror descent update. 
To see why, notice that the right-hand
side involves only quantities that are \emph{observable} at the end of
round $t$, and assuming that $\psi$ is $1$-strongly convex with respect to $\norm{\cdot}$,
we can bound each term of the second summation using Fenchel-Young inequality: 
\begin{align*}
    \inner{\gs-\hs, \ws-\wtpp}
    \le
    \cc{s}{\tpp} D_\psi(\wtpp| \ws)
    + \frac{\norm{\gs-\hs}^{2}_{*}}{2\cc{s}{\tpp}},
\end{align*}
for any choice of weights $\cc{s}{\tpp}$
satisfying $\cc{s}{\tpp}>0$ for all $s\in \obs_t$ and $t\in[T]$.
Hence, following \Cref{eq:feedback-decomp} we can further upper-bound $\sumtT \inner{\gt,\wt-\cmp}$ by
\begin{align}
    \sumtT
    \Bigg[
        \Inner{
            \htpp+\sum_{s\in\obs_t}(\gs-\hs),
            \wtpp-\cmp
        }
        +
        \sum_{s\in\obs_t}
        \cc{s}{\tpp}D_\psi(\wtpp\mid\ws)
    \Bigg]
    +\sumtT\sum_{s\in\obs_t}
    \frac{\norm{\gs-\hs}_*^2}{2\cc{s}{\tpp}}.
    \label{eq:motivating-bound}
\end{align}
This naturally suggests choosing $w_{t+1}$ to minimize the
expression inside the brackets, resulting in a mirror descent  step of the form
\begin{align}
    \wtpp \in \argmin_{\w\in\cW}
  \Set{
        \Inner{
            \htpp+\sum_{s\in\obs_t}(\gs-\hs),
            \w
        }
        +
        \sum_{s\in\obs_t}
        \cc{s}{\tpp}D_\psi(\w\mid\ws)
  }\;.
    \label{eq:simple-mao}
\end{align}
In words, the regret decomposition suggests regularizing
toward the past iterates corresponding to the gradients that
have been newly-observed on round $t$.

However, while \Cref{eq:simple-mao} naturally controls the
delayed prediction-error terms, our goal is not merely to
obtain sublinear regret under arbitrary delays.
Rather, we seek a best-of-both-worlds guarantee that
recovers the optimal worst-case dependence on delays
while retaining the quadratic prediction-error dependence
and negative stability terms needed for the fast rates obtained in \Cref{sec:main-results}. 
Achieving this requires controlling the accumulation of
Bregman-divergence terms across rounds.
In particular, unlike standard
mirror descent, the use of multiple anchors introduces a penalty
of the form
\begin{align*}
    \sumtT\sum_{s\in\obs_t}
    \cc{s}{\tpp}
    \sbrac{
        D_\psi(\cmp\mid\ws)
        -
        D_\psi(\cmp\mid\wtpp)
  },
\end{align*}
which \emph{does not telescope in the usual way} (see \Cref{sec:analysis} for a more detailed discussion).
Controlling this penalty requires a careful assignment of
regularization weights across rounds, as
restricting regularization to only the newly observed gradients prevents the cancellation of Bregman-divergence terms associated with rounds without feedback, potentially causing the above penalty to accumulate over time. 

This motivates a more general update in which
\emph{any previous iterate can serve as an anchor}.
Specifically, we introduce
\emph{Multi-Anchor Optimistic Online Mirror Descent} (\MAO),
defined by
\begin{align}
    \tag{MAO}\label{eq:update}
    w_{t+1}\in\argmin_{w\in\cW}
    \left\{
        \Inner{
            \htpp+\sum_{s\in\obs_t}(\gs-\hs),
            w
        }
        +
        \sum_{s=0}^{t}
        \cc{s}{t+1}D_\psi(w\mid w_s)
    \right\},
\end{align}
where $w_0\in\cW$ is a fixed initial anchor and the
weights $\cc{s}{t+1}$ are non-negative.
The additional flexibility in \MAO allows us to separate
two roles of the regularization weights: controlling the
discrepancies $w_s-w_{t+1}$ arising from newly observed
gradients, and managing the accumulation of
Bregman-divergence terms across rounds.
Indeed, combining \Cref{eq:feedback-decomp} with the
standard mirror descent three-point inequality yields
\begin{align}
    \sumtT \inner{\gt,\wt-\cmp}
    &\le
    \underbrace{
        \sumtT\sum_{s=0}^{t}
        \cc{s}{\tpp}
        \left[
            D_\psi(\cmp\mid\ws)
            -
            D_\psi(\cmp\mid\wtpp)
        \right]
    }_{\text{Multi-anchor potential}}
    +
        \underbrace{\sumtT\sum_{s\in\obs_t}
        \frac{\norm{\gs-\hs}_*^2}{2\cc{s}{\tpp}}}_{\text{Prediction-error}},
    \label{eq:mao-motivating-regret}
\end{align}
provided that the weights corresponding to newly observed gradients are positive. 
We formalize these intuitions in \Cref{sec:analysis}, where we establish the general regret guarantee of \MAO in \Cref{thm:general-mao-delays}.

Our main technical contribution is a
\emph{weight-allocation strategy} 
for choosing the weights
$\cc{s}{t+1}$ that controls the non-telescoping multi-anchor potential
while balancing the prediction-error cost.
As we show in \Cref{sec:algorithm}, this leads to regret
guarantees that recover the minimax-optimal
$\order(\sqrt{T+\dtot})$ dependence on delays,
while exploiting predictable feedback to obtain sharper bounds in smooth games.
In particular, when both players employ \MAO,
our analysis yields an $\order(\qmax)$ social-regret
guarantee.

\subsection{Anchor weights for delayed feedback}
\label{sec:algorithm}\label{sec:anchor-weights}

We now instantiate the multi-anchor update \eqref{eq:update} for delayed feedback.
Our goal is to choose the regularization weights so that the multi-anchor potential in \Cref{eq:mao-motivating-regret} remains controlled, while retaining sufficiently large weights for the anchors corresponding to newly observed gradients.

To understand the constraints on these weights, define $\inflow_{t}\defeq\sum_{k=0}^{t-1}\cc{k}{t}$ and $\outflow_{t}\defeq\sum_{k=t}^{T}\cc{t}{k+1}$.
Here, $\inflow_{t}$ is the total regularization weight used to compute $w_t$,whereas $\outflow_{t}$ is the total weight assigned to $w_t$ as an anchor in subsequent updates.
Rearranging the multi-anchor potential in \Cref{eq:mao-motivating-regret}, and dropping a nonpositive terminal term, gives
\begin{align*}
    \sumtT\sum_{s=0}^{t}\cc{s}{\tpp}\sbrac{D_{\psi}(\cmp|\ws)-D_{\psi}(\cmp|\wtpp)}
    &\le
    \outflow_{0}D_\psi(\cmp\mid \w_{0}) + \sumtT \bigl(\outflow_{t}-\inflow_{t}\bigr) D_\psi(u\mid \wt) \;.
\end{align*}
Thus, to control this potential, it suffices to ensure that $\outflow_t \le \inflow_t$ for every $t\in[T]$.
In other words, each iterate can be used as an anchor with total weight at most that assigned to it when it was computed.

The fixed reference point $w_0$ plays a special role in this construction.
Rather than being associated with any delayed gradient, it serves as an auxiliary anchor through which additional regularization weight can be introduced adaptively.
Indeed, under the above constraints, the only remaining positive contribution to the multi-anchor potential is $\outflow_{0}D_\psi(u\mid w_0)\le M \outflow_{0}$.
This allows us to adaptively increase the available regularization weight as needed, without advance knowledge of the delays or prediction errors, provided that the total weight $\outflow_0$ assigned to $w_0$ can be kept sufficiently small.

It remains to determine how the available weight should be distributed among the other anchors.
The prediction-error term in \Cref{eq:mao-motivating-regret} favors assigning a large weight to $\ws$ when its gradient $\gs$ arrives.
We therefore seek to preserve as much of the weight associated with each iterate $\ws$ as possible until its feedback is observed, so as to set $\cc{s}{s+d_s+1}$ sufficiently large while allowing portions of the weight available for $\ws$ to be used in earlier updates.
To formalize this idea, we introduce an \emph{allocation schedule}, which specifies the fraction of regularization weight retained by each iterate over time.

\begin{definition}[Allocation schedule] \label{def:schedule}
    A nonnegative array $(\ccdel{k}{t})_{1\le k\le t\le T+1}$ is an \emph{allocation schedule} if it satisfies the following three properties for all $t\in[T]$:
    \begin{enumerate}[label=\upshape(S\arabic*),ref=S\arabic*,leftmargin=15mm,itemsep=0pt]
        \item\label{S1}\makebox[7em][l]{monotonicity:} $\ccdel{t}{k} \ge \ccdel{t}{k+1}$ for all $k=t,\dots,T$,
        \item\label{S2}\makebox[7em][l]{zero tail:} $\ccdel{t}{k} = 0$ for all $k \ge t+d_t+1$,
        \item\label{S3}\makebox[7em][l]{capacity:} ${\sum_{k=1}^{t}} \ccdel{k}{t} = {\sum_{k\in\act_t}} \ccdel{k}{t} \le 1$. %
    \end{enumerate}
\end{definition}

Given any allocation schedule, we also introduce the notion of \emph{remaining weight} $\ccrem{s}{t}$ for some anchor $w_s$ at time $t$.
In particular, given $\gamma > 0$ and any nondecreasing sequence $0 = \eta_1 \le \dots \le \eta_{T+1}$, we define $\ccrem{s}{t} \defeq \brb{\gamma + \eta_s\ccdel{s}{t}} \Ind\{s \in \act_t\}$ for every $1 \le s \le t \le T+1$ with $s \le T$; note that we add $\gamma$ to guarantee strict positivity even when $\eta_s\ccdel{s}{t} = 0$.

Further let $Q_t \defeq \max_{s \in [t]} q_s$ for every $t \in [T]$ with $Q_0 = 0$.
We then introduce $\beta_t \defeq \eta_t + \gamma Q_t$ as an inflated version of $\eta_t$ to compensate for the incremental updates of $\eta_t$ with no prior knowledge while simultaneously preserving monotonicity.
Then, $\beta_t$ effectively represents the amount of weight flowing through the subset $[t]$; i.e., it corresponds to the in-flow from the source $0$ to $[t]$ and, by the conservation of flow from enforcing $\inflow_t = \outflow_t$, to the out-flow from $[t]$ to $\{t+1, \dots, T+1\}$.
To realize this flow, we define the regularization coefficients by releasing the weight no longer reserved at each anchor $w_s$, assigning the unallocated remainder $\ccrem{s}{t}-\ccrem{s}{t+1}$ to $w_{t+1}$, and injecting the increment $\beta_{t+1}-\beta_t$ through $w_0$:
\begin{equation} \label{eq:coefficients}
    \cc{0}{t+1} \defeq \beta_{t+1} - \beta_t \;,
    \qquad
    \cc{s}{t+1} \defeq \bbrb{\beta_t - \sum_{k \in \act_t} \ccrem{k}{t}} \Ind\{s=t\} + \ccrem{s}{t} - \ccrem{s}{t+1} \;,
    \quad \forall s \in [t] \;,
\end{equation}
where we define $b_t \defeq \beta_t - \sum_{k \in \act_t} \ccrem{k}{t}$ as the remainder weight at $t$ which we use on the anchor $w_t$ when computing $w_{t+1}$ to satisfy the corresponding flow-conservation constraint with respect to the subset $[t]$.
Note that the extra weight $\beta_{t+1} - \beta_t = \eta_{t+1} - \eta_t + \gamma(Q_{t+1} - Q_t)$ added from the default anchor $w_0$ into $w_{t+1}$ allows to adaptively increase the flow value (coming from the source $w_0$) from $\beta_t$ to $\beta_{t+1}$, applying the correction with respect to $\eta_{t+1}$ and $Q_{t+1}$ due to the newly observed feedback.

Finally, to obtain the negative stability terms needed for fast regret rates in
games, we introduce an additional anchor at the iterate $w_{s_{t+1}}$ used to form the optimistic hint.
Specifically, we add the regularization term $\gamma D_\psi(w\mid w_{s_{t+1}})$ to the update.
Unlike the reference anchor $w_0$, whose role is to supply additional regularization weight, this anchor relates the new iterate to the action at which the predicted gradient $h_{t+1} = g_{s_{t+1}}$ was computed.
Its contribution yields the negative movement term $-\frac{\gamma}{2}\norm{w_\tpp - w_{s_{t+1}}}^2$ as required by \Cref{thm:main-optimistic}, while the additional comparator-divergence terms can be controlled using the fact that each observed gradient is selected as a hint at most once.
While this always fits within our proposed multi-anchor OMD framework (by inflating each coefficient $\cc{s_t}{t}$ by $\gamma$), we treat it separately for convenience.

All these design choices lead to the final definition of \MAO
in \Cref{alg:mao}. 

\begin{restatable}{theorem}{MasterBound}\label{thm:master}
    \emph{(Master bound)}
    Let $\psi$ be differentiable and $1$-strongly convex with respect to $\norm{\cdot}$, with $\sup_{u \in \cW} D_\psi(u\mid w_0)\le M$ for every $u\in\cW$.
    Let $\gamma>0$, let $(\ccdel{k}{t})_{1\le k\le t\le T+1}$ be an allocation schedule, and let $0 = \eta_1 \le \dots \le \eta_{T+1}$.
    Then \cref{alg:mao} guarantees, for every $u\in\cW$,
    \begin{equation*}
        \sumtT\inner{g_t,w_t-u}
        \le
        2\gamma M\qmax + M\eta_{T+1} + \sumtT\frac{\dnorm{g_t-h_t}^2}{\gamma + \eta_t\ccdel{t}{t+d_t}} - \frac{\gamma}{4} \sumtT \norm*{w_{t+1} - w_{s_{t+1}}}^2 \;.
    \end{equation*}
\end{restatable}

We can perform an adaptive tuning of $\eta_t$ using a technique in close resemblance to the recurrence proved for the sequence $(\Delta_t)_{t\ge1}$ of the \textsc{AdaFTRL} algorithm \cite[Lemmas~6 and~7]{orabona2018scalefree}, or to the Stability-Penalty-Matching update rule for the sequence $(\beta_t)_{t\ge1}$ from \cite{ito2024adaptive}.
In particular, we can rewrite the penalty term from \Cref{thm:master} as a telescoping sum $M\eta_{T+1} = M\sum_t(\eta_{t+1}-\eta_t)$ and the stability term as $\sum_t\frac{\dnorm{g_t-h_t}^2}{\gamma+\eta_t\ccdel{t}{t+d_t}} = \sum_t\sum_{s\in\obs_t}\frac{\dnorm{g_s-h_s}^2}{\gamma+\eta_s\ccdel{s}{t}}$ (by \Cref{fact:variance-sum-swap} in the appendix).
Then, one may combine them as
\begin{equation*}
    M\eta_{T+1} + \sumtT\frac{\dnorm{g_t-h_t}^2}{\gamma+\eta_t\ccdel{t}{t+d_t}}
    = \sumtT\lrb{M(\eta_{t+1}-\eta_t) + \sum_{s\in\obs_t}\frac{\dnorm{g_s-h_s}^2}{\gamma+\eta_s\ccdel{s}{t}}}
\end{equation*}
and optimize term-by-term (via equalization) the resulting sum to lead to the incremental update of the form
\begin{equation} \label{eq:learning-rate-tuning}
    \eta_{t+1} = \eta_t + \frac{1}{M}\sum_{s\in\obs_t}\frac{\dnorm{g_s-h_s}^2}{\gamma+\eta_s\ccdel{s}{t}} \;,
\end{equation}
which only uses information observed by the learner at the end of round $t$.
This tuning then works as desired by \Cref{lem:learning-rate-sum} in the appendix, provided $\gamma\ge2G/\sqrt M$ and the allocation schedule guarantees sufficiently small values of $\ccdel{t}{t+d_t}$.

We now define the last missing piece for a concrete instantiation of \MAO for delayed feedback: a concrete allocation schedule meeting all the criteria mentioned so far.
To do so, we make use of the delay-related notion of \emph{congestion} introduced in \Cref{sec:prelim}.
We can define our allocation schedule $(\ccdel{s}{t})_{s\le t}$ via the congestion parameters $\kappa_{s,t}$ as follows:
\begin{equation} \label{eq:allocation}
    \ccdel{s}{t} \defeq \frac{\Ind\{s \in \act_t\}}{\kappa_{s,t}}
    \qquad
    \forall 1\le s\le t\le T\;.
\end{equation}
This schedule satisfies the conditions of \Cref{def:schedule}: the allocated fractions are nonincreasing, vanish once feedback arrives, and sum to at most one since $\kappa_{s,t}\ge q_t$ for every $s\in\cA_{t}$.
These choices together yield the result of \Cref{thm:main-optimistic}, restated here with complete details.
\begin{algorithm2e}[t]
\caption{\MAO for delayed feedback}
\label{alg:mao}
\KwInput{$w_0\in\cW$, regularizer $\psi$, optimism weight $\gamma>0$, allocation schedule $(\ccdel{s}{t})_{s\le t}$, learning rates $0=\eta_1\le\dots\le\eta_{T+1}$}
\KwInit{$w_1=w_0$, $s_1=0$, $h_1=0$, $g_0=0$.}
\For{$t=1,\dots,T$}{
    Play $w_t$ and observe $g_k$ for every $k\in\obs_t$.\;
    Set $s_{t+1}=\min\cO_{t+1}\setminus\{s_1,\ldots,s_t\}$ ($0$ if empty) and $h_{t+1}=g_{s_{t+1}}$ ($0$ if $t=T$).\;
    Compute $\eta_{t+1}$ 
    and $\alpha^{(t+1)}_k$ for $k\le t$, and set $\cc{k}{t+1}$ for $0\le k\le t$ as in \eqref{eq:coefficients}.\;
    Set $\ell_t = \htpp+\sum_{s\in\obs_t}(\gs-\hs)$.\;
    Update $w_{t+1}\in\argmin_{w\in\cW}\Bcb{\Inner{\ell_t,w}+\sum_{s=0}^{t}\cc{s}{t+1}D_\psi(w\mid w_s) + \gamma D_\psi(w \mid w_{s_{t+1}})}$.\; 
}
\end{algorithm2e}

\begin{manualtheorem}{\ref{thm:main-optimistic}} \label{thm:main-optimistic-restated}
    Let \Cref{ass:regularity} hold
    and let $\psi\colon \cW \to \R$ and $w_0 \in \cW$ satisfy \Cref{ass:diam}.
    Consider the allocation schedule $(\ccdel{s}{t})_{s\le t}$ in \eqref{eq:allocation} and the sequence $0=\eta_1\le\dots\le\eta_{T+1}$ following \eqref{eq:learning-rate-tuning}.
    Then \Cref{alg:mao} with $\gamma = \max\{4L, 2G/\sqrt{M}\}$ guarantees, for every $u \in \cW$,
    \begin{equation*}
        \sumtT \inner{g_t, w_t-u}
        \le
        \brb{8LM + 4G\sqrt M}\qmax + 2\sqrt{2M\sumtT \kappa_t\dnorm{g_t-h_t}^2} - L\sumtT \norm{w_t-w_{s_t}}^2 \;.
    \end{equation*}
\end{manualtheorem}

\section{Regret analysis of \MAO}
\label{sec:analysis}

In this section, we develop the regret analysis of \MAO, culminating in the
proof of \Cref{thm:main-optimistic}. We proceed in three steps. First, in
\Cref{sec:general-multi-anchor}, we establish a general regret bound for mirror
descent updates with multiple anchors, without imposing any particular structure
on the feedback. Next, in \Cref{sec:optimistic-delayed}, we combine this bound
with the delays-to-feedback decomposition of \Cref{lem:regret-decomp} to derive
an optimistic regret guarantee under arbitrary delays
\Cref{thm:general-mao-delays}. This bound highlights the two main ingredients of
our analysis: controlling the non-telescoping Bregman-divergence terms arising
from multiple anchors, while retaining the negative stability terms needed for
fast regret rates. Finally, in \Cref{sec:main-master}, we
show how the regularization weights and hint-selection rule of \MAO control
these contributions, establishing the master bound of \Cref{thm:master} and,
ultimately, the guarantee of \Cref{thm:main-optimistic}
in \Cref{sec:main-optimistic}.

\subsection{Results for multi-anchor online mirror descent}\label{sec:general-multi-anchor}

The following lemma generalizes the usual three-point mirror descent inequality to the multi-anchor case.
\begin{lemma}
\label{lem:multi-anchor-one-step}
    Let $\ell \in \R^n$, let $z_1, \dots, z_k \in \cW$, and let $\alpha_1, \dots, \alpha_k \ge 0$.
    Let $\psi: \cW \to \R$ be a convex differentiable function.
    Define $A \defeq \sum_{i=1}^k \alpha_i$ and
    \begin{equation*}
        w \in \argmin_{x \in \cW} \lcb{\inner{\ell, x} + \sum_{i=1}^k \alpha_i D_\psi(x \mid z_i)} \;.
    \end{equation*}
    Then, for every $u \in \cW$,
    \begin{equation*}
        \inner{\ell, w-u}
        \le
        \sum_{i=1}^k \alpha_i \brb{D_\psi(u \mid z_i) - D_\psi(w \mid z_i)} - A D_\psi(u \mid w) \;.
    \end{equation*}
\end{lemma}
\begin{proof}
    Fix any $u \in \cW$. Since $\cW$ is convex and $w$ minimizes the differentiable objective, the first-order optimality condition gives
    \begin{equation*}
        \inner*{\ell + \sum_{i=1}^k \alpha_i\brb{\nabla\psi(w) - \nabla\psi(z_i)}, w-u} \le 0 \;.
    \end{equation*}
    Rearranging and using the three-point identity for Bregman divergences $\inner{\grad \psi(z)-\grad \psi(w), w-u}=D_\psi(u|z)-D_\psi(u|w)-D_\psi(w|z)$  yields
    \begin{align*}
        \inner*{\ell, w-u}
        &\le
        \inner*{\sum_{i=1}^k \alpha_i \brb{\nabla\psi(z_i) - \nabla\psi(w)}, w-u}
        \\ &=
        \sum_{i=1}^k \alpha_i \inner*{\nabla\psi(z_i) - \nabla\psi(w), w-u}
        \\ &=
        \sum_{i=1}^k \alpha_i \brb{D_\psi(u \mid z_i) - D_\psi(u \mid w) - D_\psi(w \mid z_i)}
        \\ &=
        \sum_{i=1}^k \alpha_i \brb{D_\psi(u \mid z_i) - D_\psi(w \mid z_i)} - A D_\psi(u \mid w) \;,
    \end{align*}
    proving the claim.
\end{proof}

The following proposition bounds the one-step-ahead
quantity $\sumtT\inner{\ell_t,w_{t+1}-u}$, where $w_{t+1}$
is computed using $\ell_t$. We refer to this as a
\emph{cheating regret bound}.
For \MAO, we take
$\ell_t=h_{t+1}+\sum_{s\in\obs_t}(g_s-h_s)$
and apply \Cref{lem:regret-decomp} to recover a bound on
the linearized regret of the played actions.
The correction term in that decomposition yields the
prediction-error terms involving $\dnorm{g_s-h_s}^2$.

\begin{proposition}[Cheating regret] \label{prop:mao-omd-generic-bound}
    Fix $T \in \N_+$ and $w_0 \in \cW$.
    Let $\ell_0, \dots, \ell_T \in \R^n$ with $\ell_0 = 0$ and, for each $t \in \{0,\dots,T\}$ and each $0 \le k \le t$, let $\cc{k}{t+1} \ge 0$.
    For every $t \in \{0, \dots, T\}$ define
    \begin{equation*}
        w_{t+1} \in \argmin_{w\in\cW} \lcb{\inner{\ell_t,w} + \sum_{k=0}^{t} \cc{k}{t+1}D_\psi(w\mid w_k)} \;.
    \end{equation*}
    Also define $\inflow_t\defeq\sum_{k=0}^{t-1}\cc{k}{t}$ and $\outflow_t\defeq\sum_{k=t}^{T}\cc{t}{k+1}$ for each $t \in \{0,\dots,T+1\}$, with empty sums equal to zero.
    Then, for every $u \in \cW$,
    \begin{align*}
        \sumtT \inner{\ell_t,w_{t+1}-u}
        &\le
        \outflow_0 D_\psi(u\mid w_0) - \inflow_{T+1} D_\psi(u \mid w_{T+1}) + \sumtT \brb{\outflow_t-\inflow_t} D_\psi(u\mid w_t)
        \\ &\quad
        -\sum_{t=0}^T \sum_{k=0}^{t} \cc{k}{t+1} D_\psi(w_{t+1}\mid w_k) \;.
    \end{align*}
\end{proposition}
\begin{proof}
    Fix any $u \in \cW$.
    For each $t \in \{0, \dots, T\}$, apply \Cref{lem:multi-anchor-one-step} to the update defining $w_{t+1}$, with loss vector $\ell_t$, anchors $(w_k)_{k=0}^{t}$, and weights $(\cc{k}{t+1})_{k=0}^{t}$.
    Since the weights sum to $\inflow_{t+1}$, we obtain
    \begin{equation} \label{eq:cheating-inequality}
        \inner{\ell_t,w_{t+1}-u}
        \le
        \sum_{k=0}^{t} \cc{k}{t+1} \brb{D_\psi(u\mid w_k) - D_\psi(w_{t+1}\mid w_k)} - \inflow_{t+1} D_\psi(u\mid w_{t+1}) \;.
    \end{equation}
    When these inequalities are summed over $t=0,\dots,T$, the anchor $w_k$ contributes a comparator term at each later update $t=k,\dots,T$.
    Its total coefficient is therefore $\sum_{t=k}^{T}\cc{k}{t+1} = \outflow_k$.
    More precisely, exchanging the order of summation gives
    \begin{equation}
        \label{eq:cheating-rearrange-out}
        \sum_{t=0}^T \sum_{k=0}^{t} \cc{k}{t+1} D_\psi(u\mid w_k)
        =
        \sum_{k=0}^{T} \lrb{\sum_{t=k}^{T} \cc{k}{t+1}} D_\psi(u\mid w_k)
        =
        \outflow_0 D_\psi(u\mid w_0) + \sumtT \outflow_t D_\psi(u\mid w_t) \,.
    \end{equation}
    Substituting this identity into the sum of the one-step inequalities from \Cref{eq:cheating-inequality} and collecting the comparator terms yields
    \begin{align*}
        \sumtT \inner{\ell_t,w_{t+1}-u}
        &=
        \sum_{t=0}^T \inner{\ell_t,w_{t+1}-u}
        && \text{($\ell_0=0$)}
        \\ &\le
        \sum_{t=0}^T \sum_{k=0}^{t} \cc{k}{t+1} D_\psi(u\mid w_k) - \sum_{t=0}^T \inflow_{t+1} D_\psi(u\mid w_{t+1})
        && \text{(by \eqref{eq:cheating-inequality})}
        \\ &\quad
        -\sum_{t=0}^T \sum_{k=0}^{t} \cc{k}{t+1} D_\psi(w_{t+1}\mid w_k)
        \\ &=
        \outflow_0 D_\psi(u\mid w_0) + \sum_{t=1}^{T} \outflow_t D_\psi(u\mid w_t) - \sum_{t=0}^T \inflow_{t+1} D_\psi(u\mid w_{t+1})
        && \text{(by \eqref{eq:cheating-rearrange-out})}
        \\ &\quad
        -\sum_{t=0}^T \sum_{k=0}^{t} \cc{k}{t+1} D_\psi(w_{t+1}\mid w_k)
        \\ &=
        \outflow_0D_\psi(u\mid w_0) - \inflow_{T+1} D_\psi(u \mid w_{T+1}) + \sumtT \brb{\outflow_t-\inflow_t}D_\psi(u\mid w_t)
        \\ &\quad
        -\sum_{t=0}^T \sum_{k=0}^{t}\cc{k}{t+1}D_\psi(w_{t+1}\mid w_k) \;,
    \end{align*}
    This is the claimed bound.
\end{proof}

\subsection{Optimistic regret bound under delayed feedback}\label{sec:optimistic-delayed}

To apply \Cref{prop:mao-omd-generic-bound} to delayed feedback, we first restate the regret decomposition from \Cref{lem:regret-decomp} and provide its proof.
The lemma is restated below for convenience.

\regretdecomp*
\begin{proof}[Proof of \Cref{lem:regret-decomp}]
    Fix any $u \in \cW$ and any sequence $h_1, \dots, h_{T+1} \in \R^n$ with $h_1 = h_{T+1} = 0$.
    By adding and subtracting $\inner{h_t, w_t-u}$ for each $t \in [T]$, one obtains
    \begin{align*}
        \sumtT \inner{g_t, w_t-u}
        &=
        \sumtT \inner{h_t, w_t-u} + \sumtT \inner{g_t-h_t, w_t-u}
        \\ &=
        \sumtT \inner{h_{t+1}, w_{t+1}-u} + \sumtT \sum_{s \in \obs_t} \inner{g_s-h_s, w_s-u} \;,
    \end{align*}
    where the second identity uses the fact that $h_1 = h_{T+1} = 0$ and rearranges the terms of the second sum by permuting and grouping them according to their feedback arrival time.
    Focusing on the second sum, we can further add and subtract $\sum_{s \in \obs_t} \inner{g_s-h_s, w_{t+1}-u}$ for each $t \in [T]$ to obtain
    \begin{equation*}
        \sumtT \sum_{s \in \obs_t} \inner{g_s-h_s, w_s-u}
        =
        \sumtT \inner*{\sum_{s \in \obs_t} (g_s-h_s), w_{t+1}-u} + \sumtT \sum_{s \in \obs_t} \inner{g_s-h_s, w_s-w_{t+1}}
    \end{equation*}
    Combining the two math displays
    proves the claim.
\end{proof}

Now using the regret decomposition of \Cref{lem:regret-decomp} and the generic regret bound in \Cref{prop:mao-omd-generic-bound},
we can easily derive the following regret guarantee for the delayed feedback setting, by taking $\ell_t = h_\tpp + \sum_{s\in\obs_t}(\gs-\hs)$.
\begin{theorem}[Regret of the multi-anchor optimistic update rule]
\label{thm:general-mao-delays}
Fix hints $(h_t)_{t=1}^{T+1}$ from $\R^n$ with $h_1=h_{T+1}=0$ and anchor weights
$(\cc{k}{t+1})_{0\le k\le t\le T}$ from $[0,\infty)$ with $\cc{0}{1}>0$ and
$\cc{t}{t+d_t+1}>0$ for every $t\in[T]$, and let $w_1,\dots,w_{T+1}$ be given by
\eqref{eq:update} from $w_0\in\cW$, where $\psi:\cW\to \R_{\ge0}$
is $1$-strongly convex w.r.t.\ $\norm{\cdot}$.
Then, for every $u\in\cW$,
\begin{align}
\label{eq:abstract}
  \sumtT\inner{g_t,\,w_t-u}
  \ \le\ &\outflow_0\,\D(u\mid w_0)
  +\sum_{t=1}^{T}\brb{\outflow_t-\inflow_t}\D(u\mid w_t)
  +\sum_{t=1}^{T}\frac{\dnorm{g_t-h_t}^2}{\cc{t}{t+d_t+1}}
  \nonumber\\
  &-\half\sum_{t=0}^{T}\ \sum_{k=0}^t\cc{k}{t+1}\,\D(w_{t+1}\mid w_k)\,.
\end{align}
\end{theorem}
To understand the bound, let us recall the analysis of a standard optimistic mirror descent update of the form
$\wtpp = \argmin_{w\in\cW}\inner{\gt-h_{t}+h_{\tpp},\w}+\lambda_{\tpp}D_{\psi}(\w\mid\wt)$, applied in a standard (undelayed) OCO problem.
With $h_{1}=h_{T+1}=0$, the regret guarantee can be shown to be (see, e.g., \cite{orabona2026modern}):
\begin{align}
\sumtT \inner{\gt,\wt-\cmp}
  &\le
    \sumtT \lambda_{\tpp}D_{\psi}(\cmp\mid\wt)-\lambda_{\tpp}D_{\psi}(\cmp\mid\wtpp) + \sumtT \frac{\norm{\gt-h_{t}}^{2}_{*}}{2\lambda_{\tpp}}\nonumber\\
    &=
      \sumtT \brac{\lambda_{t}D_{\psi}(\cmp\,|\,\wt)-\lambda_{\tpp}D_{\psi}(\cmp\,|\,\wtpp)}
      +\sumtT (\lambda_{\tpp}-\lambda_{t})D_{\psi}(\cmp\,|\,\wt)+ \sumtT \frac{\norm{\gt-h_{t}}^{2}_{*}}{2\lambda_{\tpp}}\nonumber\\
  &=
      \lambda_{1}D_{\psi}(\cmp \mid w_{1})
    +\sumtT (\lambda_{\tpp}-\lambda_{t})D_{\psi}(\cmp \mid \wt)
      +\sumtT \frac{\norm{\gt-h_{t}}^{2}_{*}}{2\lambda_{\tpp}},\label{eq:ordinary-md-bound}
\end{align}
where the last line observes that the first summation telescopes as
\[
    \sumtT \lambda_t D_\psi(\cmp \mid \wt) - \lambda_\tpp D_\psi(\cmp \mid \wtpp)
    =
    \lambda_1D_\psi(\cmp \mid \w_1) - \lambda_{T+1}D_\psi(\cmp \mid \w_{T+1})
    \le
    \lambda_1D_\psi(\cmp \mid \w_1) \;.
\]
Now, assuming the uniform bound $D_\psi(u\mid w)\le M$ for all $u,w\in\mathcal{W}$, one typically chooses a sequence such as $\lambda_{t}=\frac{G}{\sqrt{M}}\sqrt{t}$, making the
remaining Bregman term telescope as well:
\begin{align*}
    \sumtT (\lambda_{\tpp}-\lambda_{t})D_{\psi}(\cmp\mid\wt)
    \le
    M\sumtT (\lambda_{\tpp}-\lambda_{t})
    =
    M(\lambda_{T+1}-\lambda_{1})\le \lambda_{T+1}M
    =
    \order\brac{G\sqrt{MT}} \;,
\end{align*}
which is minimax optimal in the undelayed setting. However, notice that the analogous
term in \Cref{thm:general-mao-delays} is instead
\begin{align*}
  \sumtT (\outflow_{t}-\inflow_{t})D_{\psi}(\cmp \mid \wt) \;,
\end{align*}
where $\outflow_{t}=\sum_{k=t}^{T}\cc{t}{k+1}$ and
$\inflow_{t}=\sum_{k=0}^{\tmm}\cc{k}{t}$, which \emph{does not necessarily
telescope}. Controlling this term is what requires
a carefully designed anchor weight schedule, such as the
one in \Cref{sec:algorithm}. On the other hand, notice that we also
obtain a variance term which is significantly more powerful
than the one in the usual approach: rather than
$\sumtT \frac{\norm{\gt-h_{t}}^{2}_{*}}{2\lambda_{\tpp}}$,
\MAO obtains an optimistic prediction-error penalty of
\begin{align*}
  \sumtT \frac{\norm{\gt-h_{t}}^{2}_{*}}{2\cc{t}{t+d_{t}+1}} \;.
\end{align*}
In particular, the coefficient $\cc{t}{t+d_{t}+1}$ is one which
is set \emph{at the future time when the feedback from round $t$
  is observed}, rather than on round $\tpp$. This is what enables
us to set the anchor weight schedule in such a way that
the desired
\begin{equation*}
    \order\lrb{\sqrt{\sumtT \kappa_{t}\norm{\gt-h_{t}}^{2}_{*}}}
    = \order\brac{\sqrt{\min\lcb{T+\dtot,\, \qmax \sumtT \norm{\gt-h_{t}}^{2}_{*}}}}
\end{equation*}
bound is obtained, which is key to our main results
in \Cref{sec:main-results}.

\begin{proof}[Proof of \Cref{thm:general-mao-delays}]
  The proof follows the argument sketched in \Cref{sec:mao_omd}.
  Apply \Cref{lem:regret-decomp} followed by
  \Cref{prop:mao-omd-generic-bound} to get
  \begin{align*}
        \sumtT \inner{g_t, w_t - u}
        &=
    \sumtT \inner*{h_{t+1} + \sum_{s \in \obs_t} (g_s - h_s), w_{t+1} - u}
    +\sumtT \sum_{s\in\obs_{t}}\Inner{\gs-\hs,\ws-\wtpp}\\
    &\le
        \outflow_0D_\psi(u\mid w_0) - \inflow_{T+1} D_\psi(u \mid w_{T+1}) + \sumtT \brb{\outflow_t-\inflow_t}D_\psi(u\mid w_t)
        \\ &\qquad
             -\sum_{t=0}^T \sum_{k=0}^{t}\cc{k}{t+1}D_\psi(w_{t+1}\mid w_k)
             +\sumtT \sum_{s\in\obs_{t}}\inner{\gs-\hs,\ws-\wtpp}\\
    &\le
        \outflow_0D_\psi(u\mid w_0) - \inflow_{T+1} D_\psi(u \mid w_{T+1}) + \sumtT \brb{\outflow_t-\inflow_t}D_\psi(u\mid w_t)
        \\ &\qquad
             -\half\sum_{t=0}^T \sum_{k=0}^{t}\cc{k}{t+1}D_\psi(w_{t+1}\mid w_k)\\
    &\qquad
             +\underbrace{\sumtT \sum_{s\in\obs_{t}}\inner{\gs-\hs,\ws-\wtpp}-\half\cc{s}{\tpp}D_{\psi}(\wtpp\mid\ws)}_{\eqdef(*)} \;,
    \end{align*}
    where the last line
    drops a non-positive factor 
    $-\half\sum_{t=0}^T\sum_{k\in \Set{0,\ldots,t}\setminus\obs_t}\cc{s}{\tpp}D_\psi(\wtpp\mid\ws)\le0$.
    Now by $1$-strong convexity of $\psi$ with respect to $\norm{\cdot}$,
    we have $D_{\psi}(x\mid y)\ge \half \norm{x-y}^{2}$ for $x,y\in\mathcal{W}$,
    so the terms
    in the last line can be bound as
    \begin{align*}
      (*)&\le \sumtT\sum_{s\in\obs_{t}}\frac{\norm{\gs-\hs}^{2}_{*}}{\cc{s}{\tpp}}
      +\frac{1}{4} \cc{s}{\tpp}\norm{\wtpp-\ws}^{2} - \half\cc{s}{\tpp}D_{\psi}(\wtpp\mid\ws)\\
      &\le
       \sumtT\sum_{s\in\obs_{t}}\frac{\norm{\gs-\hs}^{2}_{*}}{\cc{s}{\tpp}}+\frac14\cc{s}{\tpp}\norm{\wtpp-\ws}^2 - \frac14\cc{s}{\tpp}\norm{\wtpp-\ws}^2\\
        &=
        \sumtT \frac{\norm{\gt-h_{t}}^{2}_{*}}{\cc{t}{t+d_{t}+1}} \;,
    \end{align*}
    where the final equality uses \Cref{fact:variance-sum-swap}.
    Hence, plugging this back in above and bounding the remaining divergence with respect to $\w_{T+1}$ by
    $\inflow_{T+1}D_{\psi}(\cmp\mid\w_{T+1})\ge 0$ yields
    the stated bound:
    \begin{align*}
      \sumtT \inner{\gt,\wt-\cmp}
      &\le
        \outflow_0D_\psi(u\mid w_0) + \sumtT \brb{\outflow_t-\inflow_t}D_\psi(u\mid w_t)
        +\sumtT \frac{\norm{\gt-h_{t}}^{2}_{*}}{\cc{t}{t+d_{t}+1}}\\
        &\quad
             -\half\sum_{t=0}^T \sum_{k=0}^t\cc{k}{t+1}D_\psi(w_{t+1}\mid w_k) \;. \qedhere
    \end{align*}
\end{proof}

\subsection{Proof of the Master Bound (Theorem~\ref{thm:master})}%
\label{sec:main-master}

We now turn to the proof of the master bound of \Cref{thm:master}, which 
holds for any valid allocation schedule and is the main ingredient for 
proving our main result in \Cref{thm:main-optimistic}. We first establish two
auxiliary results controlling the additional regularization toward the
hint-source iterates in \Cref{lem:queue,lem:hint-divergence-difference-bound}.
We then verify that the allocation of regularization weights introduced in
\Cref{sec:algorithm} satisfies the properties required for the general regret
bound in \Cref{thm:general-mao-delays}, yielding the master bound of
\Cref{thm:master}.

We begin by bounding the number of rounds with no available hint source.
\begin{lemma} \label{lem:queue}
    Consider any choice of $s_1, \dots, s_{T+1}$ such that, for every $t \in [T+1]$, $s_t \in \hist_t \setminus \cS_t$ whenever $\hist_t \setminus \cS_t \neq \emptyset$ and $s_t=0$ otherwise, where $\cS_t \defeq \{s_1, \dots, s_{t-1}\}$.
    Then, $\abs{\{t\in[T] : s_{t+1}=0\}} \le \sigma_{\max}$ and  $\abs{\{t\in[T] : s_{t}=0\}} \le q_{\max}$.
\end{lemma}
\begin{proof}
    Let $\cJ_t \defeq \hist_t \setminus \cS_t$ and define $\nu_t \defeq \sum_{k=1}^{t} \Ind\lcb{s_{k+1} = 0}$ as the value of the partial sum of indicators for $t \in [T]$, and $\nu_0 \defeq 0$.
    Observe that, for every $t \ge 2$,
    \begin{equation*}
        \abs{\cJ_t}
        =
        \abs{\hist_t \setminus \cS_t}
        =
        \abs{\hist_t} - \abs{\hist_t \cap \cS_t}
        =
        t-1-\sigma_t - \sum_{k=1}^{t-1} \Ind\{s_k>0\}
        =
        \sum_{k=1}^{t-1} \Ind\{s_k=0\} - \sigma_t
        =
        \nu_{t-2} - \sigma_t + 1 \;,
    \end{equation*}
    where we used $\abs{\hist_t} = t-1-\sigma_t$, $\nu_{t-2} = \sum_{k=2}^{t-1} \Ind\{s_k=0\}$, and the fact that $s_1=0$.
    Thus, for $t \in [T]$,
    \begin{equation*}
        s_{t+1} = 0
        \;\iff\;
        \cJ_{t+1} = \emptyset
        \;\iff\;
        \sigma_{t+1} = \nu_{t-1} + 1 \;.
    \end{equation*}
    Since $0 = \nu_0 \le \dots \le \nu_T$ is monotonically non-decreasing, $\nu_0=0$, and $\nu_t - \nu_{t-1} = \Ind\{s_{t+1}=0\} \in \{0,1\}$, it means that there are $\nu_T$ rounds $t \in [T]$ such that $s_{t+1} = 0$.
    Now, let $t^* = \min\{t \in [T]: \nu_t = \nu_T\}$ and note that $\nu_{t^*} = \nu_{t^*-1} + 1$, hence $s_{t^*+1}=0$.
    This means that $\sigma_{t^*+1} = \nu_{t^*-1}+1 = \nu_{t^*}$.
    Therefore,
    \begin{equation*}
        \abs{\lcb{t \in [T]: s_{t+1}=0}}
        =
        \sumtT \Ind\lcb{s_{t+1}=0}
        =
        \nu_T
        =
        \nu_{t^*}
        =
        \sigma_{t^*+1}
        \le
        \sigma_{\max} \;.
    \end{equation*}
    The second bound then comes from observing that $s_1 = 0 \neq s_{T+1}$, concluding the proof.
\end{proof}

\Cref{lem:queue} shows that the initial anchor $\w_{0}$ is selected as a hint source at most $\sigma_{\max}$ times after the first round. Since every other hint source is selected at most once, we can use this property to control the additional Bregman-divergence terms arising from regularization toward the hint-source iterates. In particular, the following lemma shows that their total contribution is bounded by $M\sigma_{\max}$.
\begin{lemma} \label{lem:hint-divergence-difference-bound}
    Let \Cref{ass:diam} hold and consider any sequence $s_1, \dots, s_{T+1}$ satisfying the assumptions of \Cref{lem:queue}.
    Then, for every $u \in \cW$,
    \(
        \sumtT \brb{D_\psi(u \mid w_{s_{t+1}}) - D_\psi(u \mid w_t)}
        \le
        M\sigma_{\max} .
    \)
\end{lemma}
\begin{proof}
    Fix any $u \in \cW$.
    Observing that the map $t \mapsto s_{t+1} \in [t]$ is injective over $\{t \in [T] : s_{t+1} > 0\}$, we can show that
    \begin{align*}
        \sumtT D_\psi(u \mid w_{s_{t+1}})
        &\le
        \abs{\{t \in [T] : s_{t+1} = 0\}} \cdot D_\psi(u \mid w_0) + \sumtT D_\psi(u \mid w_{s_{t+1}}) \Ind\{s_{t+1} > 0\}
        \\ &\le
        \abs{\{t \in [T] : s_{t+1} = 0\}} \cdot D_\psi(u \mid w_0) + \sumtT D_\psi(u \mid w_t)
        \\ &\le
        \sigma_{\max} D_\psi(u \mid w_0) + \sumtT D_\psi(u \mid w_t) \;,
    \end{align*}
    with the last inequality using \Cref{lem:queue}.
    Therefore,
    \begin{equation*}
        \sumtT \brb{D_\psi(u \mid w_{s_{t+1}}) - D_\psi(u \mid w_t)}
        =
        \sumtT D_\psi(u \mid w_{s_{t+1}}) - \sumtT D_\psi(u \mid w_t)
        \le
        \sigma_{\max} D_\psi(u \mid w_0)
        \le
        M\sigma_{\max},
    \end{equation*}
    where the last inequality follows by \Cref{ass:diam}.
\end{proof}

We now turn to the regularization coefficients defined in
\Cref{eq:coefficients}. As discussed in \Cref{sec:anchor-weights}, the main
challenge is to choose these coefficients so that they control the
prediction-error terms $\sumtT \frac{\norm{\gt-h_{t}}^{2}}{\cc{s}{\tpp}}$ without
introducing an additional penalty from the multi-anchor potential. The following
lemma establishes two key properties of our construction: the coefficients are
nonnegative (so they define valid regularization weights) and the total incoming and outgoing weights of each iterate coincide. The latter property ensures that the corresponding Bregman-divergence terms cancel in the regret bound of \Cref{thm:general-mao-delays}
\begin{lemma} \label{lem:flow-feasibility}
    Given any $\gamma > 0$, any sequence $0 = \eta_1 \le \dots \le \eta_{T+1}$, and any allocation schedule $(\ccdel{k}{t})_{1\le s\le t\le T}$, the corresponding coefficients $(\cc{s}{t+1})_{0\le s\le t\le T}$ from \Cref{eq:coefficients} satisfy:
    \begin{enumerate}[(i),itemsep=0pt,leftmargin=*]
        \item $\cc{s}{t+1} \ge 0$ for every $0 \le s \le t \le T$;
        \item $\inflow_t = \outflow_t$ for every $t \in [T]$.
    \end{enumerate}
\end{lemma}
\begin{proof}
    Recall that, for every $t \in [T]$,
    \begin{equation*}
        \beta_t = \eta_t+\gamma Q_t \;,
        \qquad
        \ccrem{s}{t} = \brb{\gamma+\eta_s\ccdel{s}{t}} \Ind\{s\in\act_t\} \;,
        \qquad
        b_t = \beta_t - \sum_{s\in\act_t}\ccrem{s}{t} \;.
    \end{equation*}
    Also recall that we assume $t + d_t \le T$ for every $t \in [T]$.
    Consequently, no round $t$ remains active at time $T+1$, and $\ccrem{t}{T+1} = 0$ for every $t \in [T]$; likewise, $\ccdel{t}{T+1}=0$ by property (S2) of \Cref{def:schedule}.
    
    We first prove~(i).
    Both $\eta_t$ and $Q_t$ are nondecreasing.
    Together with $\beta_0 = 0$, this implies that $\beta_t$ is also nondecreasing.
    Hence, \Cref{eq:coefficients} gives $\cc{0}{t+1} = \beta_{t+1}-\beta_t \ge 0$ for every $t \in \{0,\dots,T\}$.
    Next, fix $t \in [T]$.
    Using the monotonicity of $\eta_s$ and the properties of the allocation schedule, we obtain
    \begin{align*}
        \sum_{s\in\act_t}\ccrem{s}{t}
        &=
        \gamma q_t + \sum_{s\in\act_t}\eta_s\ccdel{s}{t}
        && \text{($\abs{\act_t}=q_t$)}
        \\ &\le
        \gamma q_t + \eta_t\sum_{s\in\act_t}\ccdel{s}{t}
        && \text{($\eta_s \le \eta_t$ for $s \in \act_t$)}
        \\&\le
        \gamma q_t+\eta_t
        && \text{(property~(S3) of \Cref{def:schedule})}
        \\ &\le
        \gamma Q_t+\eta_t
        && \text{($q_t \le Q_t$)}
        \\ &=
        \beta_t \;.
    \end{align*}
    Therefore $b_t \ge 0$.
    We also claim that $\ccrem{s}{t} \ge \ccrem{s}{t+1}$ for every
    $1\le s\le t\le T$.
    Indeed, property~(S2) of \Cref{def:schedule} gives $\ccdel{s}{t}=0$ whenever $s \notin \act_t$, so
    \(
        \ccrem{s}{t}
        =
        \gamma\Ind\{s\in\act_t\} + \eta_s\ccdel{s}{t} .
    \)
    For a fixed $s\le t$, round $s$ leaves the active set between times $t$ and $t+1$ exactly when $s\in\obs_t$.
    Thus
    \[
        \ccrem{s}{t}-\ccrem{s}{t+1}
        =
        \gamma\Ind\{s\in\mathcal N_t\}
        +\eta_s\bigl(\ccdel{s}{t}-\ccdel{s}{t+1}\bigr)
        \ge0,
    \]
    where the inequality follows from $\eta_s\ge0$ and property~(S1) of \Cref{def:schedule}.
    Substituting these inequalities into \Cref{eq:coefficients} yields, for all $1\le s\le t\le T$,
    \[
        \cc{s}{t+1}
        =
        b_t\Ind\{s=t\} + \ccrem{s}{t} - \ccrem{s}{t+1}
        \ge
        0 \;.
    \]
    Together with the nonnegativity of the source coefficients $\cc{0}{t+1}$, this proves~(i).
    
    To prove~(ii), we show that both the incoming and outgoing weights at each $t\in[T]$ equal $b_t + \ccrem{t}{t}$ .
    First recall their definitions:
    \[
        \inflow_t = \sum_{s=0}^{t-1}\cc{s}{t} \;,
        \qquad
        \outflow_t = \sum_{k=t}^{T}\cc{t}{k+1} \;.
    \]
    For each $t \in \{2,\dots,T\}$, \Cref{eq:coefficients} gives
    \begin{align*}
        \inflow_t
        &=
        \cc{0}{t} + \sum_{s=1}^{t-1}\cc{s}{t}
        \\ &=
        (\beta_t-\beta_{t-1}) + b_{t-1} + \sum_{s=1}^{t-1} \brb{\ccrem{s}{t-1}-\ccrem{s}{t}}
        \\ &=
        \beta_t - \sum_{s=1}^{t-1}\ccrem{s}{t}
        && \text{(definition of $b_{t-1}$)}
        \\ &=
        \beta_t - \sum_{s \in \act_t} \ccrem{s}{t} + \ccrem{t}{t}
        && \text{(definition of $\ccrem{s}{t}$)}
        \\ &=
        b_t + \ccrem{t}{t} \;.
        && \text{(definition of $b_t$)}
    \end{align*}
    For $t=1$, the same identity holds directly because $\act_1 = \{1\}$, and thus $\inflow_1 = \cc{0}{1} = \beta_1 = b_1 + \ccrem{1}{1}$.
    Finally, fix any $t\in[T]$ and sum coefficients $\cc{t}{k+1}$ using their definition from \Cref{eq:coefficients} over all $k \in \{t, \dots, T\}$ to obtain that
    \begin{align*}
        \outflow_t
        &=
        \sum_{k=t}^{T} \lrb{b_k\Ind\{k=t\} + \ccrem{t}{k} - \ccrem{t}{k+1}}
        \\ &=
        b_t + \sum_{k=t}^{T} \brb{\ccrem{t}{k}-\ccrem{t}{k+1}}
        \\ &=
        b_t + \ccrem{t}{t} - \ccrem{t}{T+1}
        \\ &=
        b_t + \ccrem{t}{t}
        && \text{($\ccrem{t}{T+1} = 0$)}
        \\ &=
        \inflow_t \;.
    \end{align*}
    This proves~(ii).
\end{proof}

We are now ready to establish the master bound of \Cref{thm:master}. The
weight-conservation property of \Cref{lem:flow-feasibility} allows us to
eliminate the non-telescoping terms in the general regret bound of
\Cref{thm:general-mao-delays}. On top of these coefficients, \Cref{alg:mao}
introduces an additional regularization term centered at the selected
hint-source iterate, with weight $\gamma$. This term provides the negative
stability contribution needed for fast regret rates in games, while
\Cref{lem:hint-divergence-difference-bound} ensures that its additional
contribution to the multi-anchor potential remains controlled. Combining these
properties yields \Cref{thm:master}, restated for convenience below.
\MasterBound*
\begin{proof}%
    Fix $u \in \cW$.
    We begin by defining the regularization coefficients used by \Cref{alg:mao}.
    For every $t \in [T]$ and every $k \in \{0, \dots, t\}$, they are
    $\hcc{k}{t+1} \defeq \cc{k}{t+1} + \gamma\Ind\{k=s_{t+1}\}$.
    Furthermore, for each $t \in [T+1]$, these coefficients induce the total incoming weight $\hinflow_t \defeq \sum_{k=0}^{t-1} \hcc{k}{t} = \inflow_t + \gamma$ and the total outgoing weight $\houtflow_{t-1} \defeq \sum_{k=t}^{T+1} \hcc{t-1}{k} = \outflow_{t-1} + \gamma\sum_{k=t}^{T+1} \Ind\{s_k=t\}$, where $\inflow_t$ and $\outflow_{t-1}$ are defined in the usual way with respect to coefficients $\cc{k}{t+1}$ only.
    Using these definitions, we can show that
    \begin{align*}
        &\houtflow_0 D_\psi(u \mid w_0) + \sumtT \brb{\houtflow_t - \hinflow_t} D_\psi(u \mid w_t)
        \\ &\quad=
        \outflow_0 D_\psi(u \mid w_0) + \sumtT \brb{\outflow_t - \inflow_t} D_\psi(u \mid w_t)
        \\ &\qquad
        +\gamma\sum_{t=0}^T \sum_{k=t}^T D_\psi(u \mid w_t)\Ind\{s_{k+1}=t\} - \gamma\sumtT D_\psi(u \mid w_t)
        \\ &\quad=
        \outflow_0 D_\psi(u \mid w_0) + \sumtT \brb{\outflow_t - \inflow_t} D_\psi(u \mid w_t) + \gamma\sum_{k=1}^T D_\psi(u \mid w_{s_{k+1}}) - \gamma\sumtT D_\psi(u \mid w_t)
        \\ &\quad=
        \outflow_0 D_\psi(u \mid w_0) + \sumtT \brb{\outflow_t - \inflow_t} D_\psi(u \mid w_t) + \gamma\sumtT \brb{D_\psi(u \mid w_{s_{t+1}}) - D_\psi(u \mid w_t)}
        \\ &\quad\le
        \outflow_0 D_\psi(u \mid w_0) + \sumtT \brb{\outflow_t - \inflow_t} D_\psi(u \mid w_t) + \gamma M\sigma_{\max} \,,
    \end{align*}
    where the last step follows by \Cref{lem:hint-divergence-difference-bound}.
    Since $s_{t+1} \in \{0, \dots, t\}$ and $\hcc{s_{t+1}}{t+1} \ge \gamma$ for every $t \in [T]$, and using the $1$-strong convexity of $\psi$, we also have that
    \begin{align*}
        \sum_{t=0}^{T}\sum_{k=0}^t \hcc{k}{t+1} D_\psi(w_{t+1} \mid w_k)
        &\ge
        \sumtT \hcc{s_{t+1}}{t+1} D_\psi(w_{t+1} \mid w_{s_{t+1}})
        \\ &\ge
        \gamma\sumtT D_\psi(w_{t+1} \mid w_{s_{t+1}})
        \\ &\ge
        \frac{\gamma}{2}\sumtT \norm*{w_{t+1} - w_{s_{t+1}}}^2 \;.
    \end{align*}

    Because the coefficients $\cc{s}{t+1}$ are nonnegative by \Cref{lem:flow-feasibility}, we can now apply \Cref{thm:general-mao-delays} and show that
    \begin{align}
        \sumtT \inner{g_t, w_t-u}
        &\le
        \houtflow_0 D_\psi(u \mid w_0) + \sumtT \brb{\houtflow_t - \hinflow_t} D_\psi(u \mid w_t) + \sumtT \frac{\dnorm{g_t-h_t}^2}{\hcc{t}{t+d_t+1}}
        \nonumber\\ &\quad
        -\frac12\sum_{t=0}^{T}\sum_{k=0}^t \hcc{k}{t+1} D_\psi(w_{t+1} \mid w_k)
        \nonumber\\ &\le
        \gamma M\sigma_{\max} + \outflow_0 D_\psi(u \mid w_0) + \sumtT \brb{\outflow_t - \inflow_t} D_\psi(u \mid w_t) + \sumtT \frac{\dnorm{g_t-h_t}^2}{\cc{t}{t+d_t+1}}
        \nonumber\\ &\quad
        -\frac{\gamma}{4}\sumtT \norm*{w_{t+1} - w_{s_{t+1}}}^2 \;.
        \label{eq:master-bound-intermediate}
    \end{align}

    Recalling the definition of $\ccrem{t}{s} = (\gamma + \eta_t\ccdel{t}{s})\Ind\{t \in \act_s\}$, and since $t \in \act_{t+d_t}$ whereas $t \notin \act_{t+d_t+1}$, the ``terminal'' coefficients $\cc{t}{t+d_t+1}$ satisfy
    \begin{equation*}
        \cc{t}{t+d_t+1}
        \ge
        \ccrem{t}{t+d_t} - \ccrem{t}{t+d_t+1}
        =
        \gamma + \eta_t\ccdel{t}{t+d_t} \;.
    \end{equation*}
    Moreover, the total outgoing weight $\outflow_0$ with respect to the default anchor $w_0$ corresponds by construction to
    \begin{equation*}
        \outflow_0
        =
        \sum_{t=1}^{T+1}\cc{0}{t}
        =
        \sum_{t=0}^T (\beta_{t+1} - \beta_t)
        =
        \beta_{T+1} - \beta_0
        =
        \eta_{T+1} + \gamma\qmax \;,
    \end{equation*}
    and thus $\outflow_0 D_\psi(u \mid w_0) \le M(\eta_{T+1} + \gamma \qmax)$.
    Applying these inequalities to the bound shown in \Cref{eq:master-bound-intermediate} finally gives
    \begin{align*}
        \sumtT \inner{g_t, w_t-u}
        &\le
        2\gamma M\qmax + M\eta_{T+1} + \sumtT \frac{\dnorm{g_t-h_t}^2}{\gamma+\eta_t\ccdel{t}{t+d_t}} - \frac{\gamma}{4}\sumtT \norm*{w_{t+1}-w_{s_{t+1}}}^2
        \\ &\quad
        +\sumtT \brb{\outflow_t - \inflow_t} D_\psi(u \mid w_t)
        \\ &=
        2\gamma M\qmax + M\eta_{T+1} + \sumtT \frac{\dnorm{g_t-h_t}^2}{\gamma+\eta_t\ccdel{t}{t+d_t}} - \frac{\gamma}{4}\sumtT \norm*{w_{t+1}-w_{s_{t+1}}}^2 \;,
    \end{align*}
    where the last step holds because $\outflow_t = \inflow_t$ by \Cref{lem:flow-feasibility}.
    This concludes the proof.
\end{proof}

\subsection{Proof of Theorem~\ref{thm:main-optimistic}}
\label{sec:main-optimistic}

\Cref{thm:master} holds for any allocation schedule satisfying
\Cref{def:schedule}. To obtain our desired dependence on the delays, we now
specialize to the congestion-based allocation schedule introduced in
\Cref{eq:allocation}.  
  The following lemma verifies that this construction indeed satisfies all the requirements of an allocation schedule.
\begin{lemma} \label{lem:allocation-kappa}
    The sequence $(\ccdel{s}{t})_{1\le s\le t\le T+1}$ from \Cref{eq:allocation} is an allocation schedule in the sense of \Cref{def:schedule}.
\end{lemma}
\begin{proof}
    First note that all values are nonnegative.
    By \Cref{fact:kappa}(a), for every $s \in [T]$, the running congestion value $\kappa_{s,t}$ w.r.t.\ round $s$ is nondecreasing while $s \in \act_t$.
    Therefore, its reciprocal is nonincreasing over these same rounds.
    After time $s+d_s$, $s \notin \act_t$ and its allocation thus becomes $0$.
    This proves~(S1) and~(S2) of \Cref{def:schedule}.

    For~(S3) from \Cref{def:schedule}, fix $t \in [T]$.
    \Cref{fact:kappa}(a) gives $\kappa_{s,t} \ge q_t \ge 1$ for every $s \in \act_t$.
    Since $\abs{\act_t} = q_t$, we obtain
    \begin{equation*}
        \sum_{s=1}^t \ccdel{s}{t}
        =
        \sum_{s \in \act_t} \frac{1}{\kappa_{s,t}}
        \le
        \frac{\abs{\act_t}}{q_t}
        =
        1 \;. \qedhere
    \end{equation*}
\end{proof}

We can now complete the proof of \Cref{thm:main-optimistic} by combining the master bound of \Cref{thm:master} with the allocation schedule \Cref{eq:allocation} and the adaptive tuning rule \Cref{eq:learning-rate-tuning}. By \Cref{lem:allocation-kappa}, the proposed schedule satisfies \Cref{def:schedule}. Moreover, its terminal weights are inversely proportional to the congestion of the corresponding rounds, allowing us to apply \Cref{lem:learning-rate-sum} to control the accumulated prediction-error terms. Finally, the choice of the optimism weight $\gamma$ ensures that the negative stability term has the coefficient required for the fast-rate analysis.
\begin{proof}[Proof of \Cref{thm:main-optimistic}]
    Fix $u \in \cW$ and define $\lambda_t \defeq \frac{1}{M}\dnorm{g_t-h_t}^2$ for every $t \in [T]$.
    The allocation schedule satisfies \Cref{def:schedule} by \Cref{lem:allocation-kappa}, and it also guarantees
    \begin{equation*}
        0
        <
        \ccdel{t}{t+d_t}
        =
        \frac{1}{\kappa_{t,t+d_t}}
        =
        \frac{1}{\kappa_t}
        \le
        \frac{1}{q_t}
    \end{equation*}
    for every $t \in [T]$, where the last inequality holds by \Cref{fact:kappa}(a).

    Since each hint $h_t$ is either an observed gradient or zero, we have $\dnorm{h_t} \le G$ and thus $\lambda_t\le4G^2/M\le\gamma^2$ by the triangle inequality.
    Applying \Cref{lem:learning-rate-sum} with $\alpha_t=\ccdel{t}{t+d_t}$ gives $\eta_{T+1}\le\sqrt{(2/M)\sumtT\kappa_t\dnorm{g_t-h_t}^2}$.
    Moreover, the chosen weight satisfies $2\gamma M\le8LM+4G\sqrt M$ and $\gamma/4\ge L$.
    Since $w_1=w_{s_1}=w_0$, dropping the nonpositive terminal movement term at $T+1$ ultimately yields:
    \begin{align*}
        \sumtT \inner{g_t, w_t-u}
        &\le
        2\gamma M\qmax + M\eta_{T+1} + \sumtT \frac{\dnorm{g_t-h_t}^2}{\gamma + \eta_t\ccdel{t}{t+d_t}} - \frac{\gamma}{4}\sumtT \norm*{w_{t+1}-w_{s_{t+1}}}^2
        \\ &=
        2\gamma M\qmax + 2M\eta_{T+1} - \frac{\gamma}{4}\sumtT \norm*{w_{t+1}-w_{s_{t+1}}}^2
        \\ &\le
        \brb{8LM + 4G\sqrt{M}}\qmax + 2\sqrt{2M\sumtT \kappa_t\dnorm{g_t-h_t}^2} - L\sumtT \norm*{w_t-w_{s_t}}^2 \;. \qedhere
    \end{align*}
\end{proof}

\section{Multiplayer games}
\label{app:multi}\label{sec:multi}

In this section, we consider $N$-player smooth games with player-independent delays, and we show that \MAO with optimism weight $\gamma = \max\{2NL,2G/\sqrt M\}$ achieves fast social regret during self-play, fast individual regret when the game has nonnegative social regret, and the optimal regret rate under delays against arbitrary opponents.

\paragraph{Setting.}
We consider the $N$-player smooth games of~\cite{anagnostides2022lastiterate} under delays.
Player $p\in[N]$ chooses an action $w^p$ in a nonempty, compact, and convex set $\cW_p$, and at a strategy profile $z=(w^1,\ldots,w^N)$, it incurs a loss $f_p(z)$ that is differentiable and convex in $w^p$.
Denote by $g^p(z)\defeq\nabla_{w^p}f_p(z)$ the loss gradient of player $p$, and by $(u,z^{-p})$ the profile $z$ with $w^p$ replaced by $u\in\cW_p$.
We assume that there exist constants $G, L > 0$ such that
\begin{equation*}
\dnorm{g^p(z)}\le G
\qquad \text{and} \qquad
\dnorm{g^p(z)-g^p(z')}
\le
L\sum_{j=1}^N\norm{w^j-w'^j}
\end{equation*}
for every player $p\in[N]$ and all profiles $z=(w^1,\ldots,w^N)$ and $z'=(w'^1,\ldots,w'^N)$, which recovers \Cref{ass:smooth} when $N=2$.
In each round $t$, the players play the profile $z_t=(w^1_t,\ldots,w^N_t)$, and the gradients $g^p_t\defeq g^p(z_t)$ of round $t$ reach all players at the end of round $t+d_t$, where the delays $d_t$ are arbitrary and unknown.
The individual regret of player $p$ is $\Reg^p_T\defeq\sup_{u\in\cW_p}\sumtT\brb{f_p(z_t)-f_p(u,z^{-p}_t)}$, and the social regret is $\SocReg_T\defeq\sum_{p=1}^N\Reg^p_T$.

\paragraph{Games with nonnegative social regret.}
In general-sum games, the social regret can be negative, so a bound on the social regret does not bound the individual regrets.
As in~\cite[Theorem~3.1 and Corollary~3.3]{anagnostides2022lastiterate}, we obtain individual bounds when the social regret is nonnegative.

\begin{assumption}[Nonnegative social regret]
\label{ass:multi-stable}
For every horizon $T \in \N_+$ and every sequence of profiles $z_1,\dots,z_T \in \prod_p \cW_p$, we have $\SocReg_T\ge0$.
\end{assumption}
\Cref{ass:multi-stable} holds in two-player zero-sum games, including smooth convex--concave games, in zero-sum and constant-sum polymatrix games~\cite{cai2016polymatrix}, and in strategically zero-sum games and their polymatrix extensions~\cite[Proposition~A.10]{anagnostides2022lastiterate}.

\paragraph{Algorithm.}
Each player $p$ runs \Cref{alg:mao} with its own gradients, a regularizer $\psi_p$ satisfying \Cref{ass:diam} with the same $M$, and optimism weight $\gamma=\max\{2NL,2G/\sqrt M\}$, which agrees with the two-player choice when $N=2$.
Since the delays are player-independent, all players select the same hint sources $s_t$.
The proof of \Cref{thm:main-optimistic} first gives a bound for any $\gamma\ge2G/\sqrt M$: applying \Cref{thm:master,lem:learning-rate-sum} with the schedule \eqref{eq:allocation} and tuning \eqref{eq:learning-rate-tuning}, and dropping the terminal movement term at $T+1$, we obtain, for every $u\in\cW$,
\begin{equation}
\label{eq:master-gamma}
\sumtT\inner{g_t^p,w_t^p-u}
\le
{2\gamma M\qmax+2\sqrt{2M\sumtT\kappa_t\dnorm{g_t^p-h_t^p}^2}-\frac{\gamma}{4}\sumtT\norm{w_t^p-w_{s_t}^p}^2} \;,
\end{equation}
{Taking the two-player parameter choice for $\gamma$ and bounding its coefficients recovers \Cref{thm:main-optimistic}.}

\begin{proposition}
\label{prop:main-multi}
If all players run \MAO with optimism weight $\gamma = \max\{2NL,2G/\sqrt M\}$ under arbitrary unknown 
delays, then $\SocReg_T=\order\brb{N(G\sqrt M+NLM)\qmax}$.
If, in addition, \Cref{ass:multi-stable} holds, then $\Reg^p_T=\order\brb{(G\sqrt M+NLM)\qmax}$ for every player $p\in[N]$.
Furthermore, a player that runs \MAO with optimism weight $\gamma = \max\{2NL,2G/\sqrt M\}$ against arbitrary opponents under arbitrary unknown delays guarantees $\Reg^p_T=\order\brb{G\sqrt{M(T+\dtot)}+NLM\qmax}$.
\end{proposition}

\begin{proof}
Denote by $\Delta_T^p\defeq\sum_{t\in[T]}\norm{w^p_t-w^p_{s_t}}^2$ the movement of player $p$ relative to the hint sources, and let $\Delta_T\defeq\sum_{p=1}^N\Delta_T^p$.
At the $\qmax$ rounds with $s_t=0$, the hint is $0$ and $\dnorm{g^p_t-h^p_t}\le G$.
At every other round, the hints $h^p_t=g^p(z_{s_t})$ of all players come from the same profile, so smoothness and the Cauchy-Schwarz inequality give
\begin{equation*}
\dnorm{g^p_t-h^p_t}^2
\le
NL^2\sum_{j=1}^N\norm{w^j_t-w^j_{s_t}}^2 \;.
\end{equation*}
Together with $\kappa_t\le\qmax$, we obtain that $\sumtT\kappa_t\dnorm{g^p_t-h^p_t}^2\le\qmax\brb{G^2\qmax+NL^2\Delta_T}$.
Since $f_p(z_t)-f_p(u,z^{-p}_t)\le\inner{g^p_t,w^p_t-u}$ by convexity, \Cref{eq:master-gamma} with $\gamma=\max\{2NL,2G/\sqrt M\}$ and $\sqrt{a+b}\le\sqrt a+\sqrt b$ give, for each player $p\in[N]$,
\begin{equation}
\label{eq:main-multi-player}
\Reg^p_T
\le
{\brb{4NLM+(4+2\sqrt2)G\sqrt M}\qmax+2\sqrt2\,L\sqrt{NM\qmax\Delta_T}-\frac{NL}{2}\Delta_T^p} \;.
\end{equation}Summing \eqref{eq:main-multi-player} over the players, we obtain that
\begin{equation}
\label{eq:main-multi-social}
\SocReg_T
\le
{N\brb{4NLM+(4+2\sqrt2)G\sqrt M}\qmax+2\sqrt2\,NL\sqrt{NM\qmax\Delta_T}-\frac{NL}{2}\Delta_T} \;.
\end{equation}
By the AM-GM inequality, {$2\sqrt2\,NL\sqrt{NM\qmax\Delta_T}\le NL\Delta_T/2+4N^2LM\qmax$}, so {$\SocReg_T\le\brb{(4+2\sqrt2)NG\sqrt M+8N^2LM}\qmax$}.

Under \Cref{ass:multi-stable}, the left-hand side of \eqref{eq:main-multi-social} is nonnegative.
Using {$2\sqrt2\,NL\sqrt{NM\qmax\Delta_T}\le NL\Delta_T/4+8N^2LM\qmax$} in \eqref{eq:main-multi-social} and rearranging the terms gives {$L\Delta_T\le\brb{(16+8\sqrt2)G\sqrt M+48NLM}\qmax$}.
Substituting this bound into \eqref{eq:main-multi-player} with {$2\sqrt2\,L\sqrt{NM\qmax\Delta_T}\le L\Delta_T/4+8NLM\qmax$} gives {$\Reg^p_T\le\brb{(8+4\sqrt2)G\sqrt M+24NLM}\qmax$}.

Against arbitrary opponents, each hint of player $p$ is an observed gradient or zero, so $\dnorm{g^p_t-h^p_t}^2\le4G^2$, and \Cref{fact:kappa}(b) gives $\sumtT\kappa_t\dnorm{g^p_t-h^p_t}^2\le4G^2(T+2\dtot)$.
Dropping the negative term in \eqref{eq:master-gamma} with {$\gamma=\max\{2NL,2G/\sqrt M\}$}, we obtain that {$\Reg^p_T\le(4NLM+4G\sqrt M)\qmax+4\sqrt2\,G\sqrt{M(T+2\dtot)}\le(8+4\sqrt2)G\sqrt{M(T+\dtot)}+4NLM\qmax$}, {where the last inequality uses $\qmax^2\le T+2\dtot$ from \Cref{fact:sets}(e)}.
\end{proof}

\section{Player-dependent delays}
\label{app:hetero}\label{sec:hetero}
In this section, we consider two-player zero-sum games in which the two players can receive the feedback of the same round at different times, and we show that \MAO keeps the guarantees of \Cref{thm:main-games} when the players communicate which feedback they have received.

\paragraph{Setting.}
We consider the two-player zero-sum game of \Cref{sec:prelim} with player-dependent delays: the gradient $g^x_t$ of round $t$ reaches player $x$ at the end of round $t+d^x_t$, and the gradient $g^y_t$ of round $t$ reaches player $y$ at the end of round $t+d^y_t$, where the delays $d^x_t,d^y_t\ge0$ are arbitrary and unknown.
For each player $p\in\{x,y\}$, denote by $\miss^p_t\defeq\{s\in[t-1]:s+d^p_s\ge t\}$ the rounds whose gradient for player $p$ is pending at the start of round $t$, and let $\sigma^p_{\max}\defeq\max_{t\in[T]}\abs{\miss^p_t}$.
At the end of each round, the players communicate the indices of the rounds whose gradients they have received so far, but not the gradients themselves.

\paragraph{Algorithm.}
With this communication, at the end of round $t$ both players know the past rounds $s$ for which both gradients have arrived, that is, the rounds with $s+\widetilde d_s\le t$ for the player-independent effective delay
\begin{equation*}
\widetilde d_s\defeq\max\{d^x_s,d^y_s\} \;.
\end{equation*}
Each player stores its gradient of round $s$ until the end of round $s+\widetilde d_s$, and only then passes it to \MAO.
Thus, both players run \Cref{alg:mao} with optimism weight $\gamma=\max\{4L,2G/\sqrt M\}$, without any modification, under the player-independent delays $(\widetilde d_t)_{t\in[T]}$.
In particular, the observed sets, the hint sources $s_t$, and the anchor weights $\cc{k}{t+1}$ are computed from $(\widetilde d_t)_{t\in[T]}$, and they are the same for both players.

\begin{theorem}
\label{thm:hetero}
Let \Cref{ass:smooth,ass:diam} hold.
Under player-dependent delays, if both players run \textnormal{\MAO} under the effective delays $\widetilde d_t=\max\{d^x_t,d^y_t\}$, then
$\SocReg_T$, $\Reg^x_T$, and $\Reg^y_T$ are all $\order\brb{(G\sqrt M+LM)(\sigma^x_{\max}+\sigma^y_{\max}+1)}$.
\end{theorem}

\begin{proof}
Denote by $\widetilde\miss_t\defeq\{s\in[t-1]:s+\widetilde d_s\ge t\}$ the set of pending rounds under the effective delays, and by $\widetilde q_{\max}$ the value of $\qmax$ under the effective delays.
Since $s+\widetilde d_s\ge t$ if and only if $s+d^x_s\ge t$ or $s+d^y_s\ge t$, we have $\widetilde\miss_t=\miss^x_t\cup\miss^y_t$, and therefore
\begin{equation*}
\widetilde q_{\max}
=
1+\max_{t\in[T]}\abs{\widetilde\miss_t}
\le
1+\max_{t\in[T]}\brb{\abs{\miss^x_t}+\abs{\miss^y_t}}
\le
\sigma^x_{\max}+\sigma^y_{\max}+1 \;.
\end{equation*}
For each round $s$, the gradients that \MAO receives are $g^x(x_s,y_s)$ and $g^y(x_s,y_s)$, and both players receive them at the end of round $s+\widetilde d_s$.
Hence, both players run \MAO under the player-independent delays $(\widetilde d_t)_{t\in[T]}$, and \Cref{thm:main-games} gives that $\SocReg_T$, $\Reg^x_T$, and $\Reg^y_T$ are all $\order\brb{(G\sqrt M+LM)\widetilde q_{\max}}$.
Substituting the bound on $\widetilde q_{\max}$ proves the theorem.
\end{proof}

Without communication, each player can still run \MAO under its own delays, and \Cref{thm:main-robust} bounds its individual regret against an arbitrary opponent.

\section{Discussion}
\label{sec:discussion}

In this paper, we studied learning in smooth games with arbitrary and unknown
feedback delays. We introduced \MAO{}, an optimistic learning algorithm that
handles delayed feedback directly through an adaptive allocation of
regularization weights across past iterates. Our algorithm achieves
optimal $O(\sigma_{\max}+1)$ social and individual regret when both players use it,
while simultaneously guaranteeing minimax-optimal $O(\sqrt{T+d_{\mathrm{tot}}})$
regret against arbitrary opponents. These guarantees improve upon
scheduling-based approaches, for which we showed that an additional logarithmic
factor cannot generally be avoided. We also extended our results to multiplayer
games and settings with player-dependent delays.

\section*{AI methodology statement}
\label{sec:ai-disclosure}
We used OpenAI's GPT-6 Astra and Anthropic's Claude Fable~5.1 models
to explore initial proof strategies, refine draft arguments, and check mathematical derivations,
including
exposition of the scheduling-based reduction and its lower-bound analysis (\Cref{sec:bold,app:assignment}), the multi-anchor regret analysis (\Cref{sec:mao_omd,sec:analysis}), and supporting lemmas (\Cref{app:auxiliary}).
The AI-assisted arguments were thoroughly revised by the authors rather than adopted as standalone results.
The authors take full responsibility for the correctness, originality, and integrity of the paper, including all mathematical arguments and references.

\phantomsection
\addcontentsline{toc}{section}{References}
\bibliographystyle{alpha-arxiv}
\bibliography{refs}

\newpage
\appendix
\crefalias{section}{appendix}
\section{Additional related work} \label{app:related-work}

\paragraph{Online learning.}
Online learning and online convex optimization have been studied extensively; see \cite{cesa2006prediction,shalev2012online,orabona2026modern} for background.
Optimistic methods, which exploit a prediction of the next loss, were analyzed via predictable sequences by \cite{chiang2012,rakhlin2013predictable}; see also \cite{joulani2017modular} for a modular treatment.
The multi-anchor regularization of \MAO is similar in spirit to ``proximal'' FTRL methods, which use a regularizer of the form $\Psi_t(w)=\sum_{s=1}^t\frac{c_s}{2}\norm{w-w_s}^2$ with a typically monotonic sequence $(c_s)_s$ \citep{mcmahan2010adaptive,mcmahan2011follow}.
There, however, the weight on a past action $w_s$ is fixed once and stays the same at every later round, whereas \MAO uses a weight $c^{(t)}_s$ on $w_s$ that changes with the round $t$, determined adaptively by the delays that unfold after round $s$ and by when the feedback of round $s$ arrives.
The adaptive tuning of the regularization scale in \MAO follows the tradition of scale-free and parameter-free methods \citep{orabona2018scalefree,jacobsen2022parameterfree,ito2024adaptive}.

\paragraph{Online learning with delayed feedback.}
Early work on delayed feedback includes \cite{weinberger2002}, who showed that $d+1$ interleaved copies of a base learner are optimal under a constant delay $d$, and the black-box reduction BOLD of \cite{joulani2013} for arbitrary delays. Beyond black-box techniques, \cite{quanrud2015} showed that gradient descent and FTPL applied directly to the delayed gradients achieve the minimax-optimal $O(\sqrt{T+\dtot})$ regret, and there is substantial work on algorithms tailored to delayed online convex optimization, including reductions to immediate feedback \citep{joulani2016delay, ryabchenko2026reduction}, dynamic regret \citep{wan2024nonstationary,esposito2026dynamic}, strongly adaptive regret \citep{jacobsen2026strongly}, curvature \citep{qiu2025curvature}, universal guarantees \citep{qiu2026universal}, and decentralized settings \citep{qiu2026decentralized}. Closest in spirit to our regret decomposition is the delay-as-optimism view of \cite{flaspohler2021}, which treats the missing gradients under a constant delay as an optimistic hint. Our weight allocation is inspired by the Banker-OMD principle of \cite{pmlr-v202-huang23e,qiu2026universal}. Delays also arise in multi-armed bandit problems, where they interact with exploration \citep{cesabianchi2019delay,thune2019,bistritz2019delayed,zimmert2020,gyorgy2021,vanderhoeven2022,masoudian2022,esposito2023delayed,schlisselberg2026linear}.  Recently, delays have also been studied at the intersection of online learning and scheduling, where the learner has a limited capacity for tracking feedback from pending rounds and must choose which feedback to receive \citep{ryabchenko2025capacity,ryabchenko2026capacityOCO}. Scheduling plays a different role there than in the BOLD reduction, which receives all feedback and only decides which learner handles each round.

\paragraph{Learning in games.}
The connection between regret minimization and equilibria in zero-sum games goes back to \cite{freund1999adaptive}; see also \cite{farina2019online,cai2016polymatrix}.
Fast convergence through optimism originates in the optimistic gradient method of \cite{popov1980} and the analyses of \cite{rakhlin2013games,syrgkanis2015fast}, whose RVU property we use in the form of a negative movement term; see \cite{daskalakis2018training,hsieh2019singlecall} for last-iterate convergence, \cite{hsieh2021adaptive} for adaptive regret and convergence in continuous games, \cite{chen2020hedging,daskalakis2021nearoptimal,anagnostides2022,anagnostides2022lastiterate} for general-sum games, and \cite{liu2026constant,abbadi2026constant,soleymani2026multiplicative,mai2026constant} for recent $T$-independent bounds.
Delays in games were first studied in continuous games \citep{zhou2017,heliou2020,hsieh2022}; fast rates under delays are recent and limited to fixed delays \citep{fujimoto2025,fujimoto2026linear,zhuang2026optimal}.

\paragraph{Online interval partitioning.}
The assignment of rounds to independent learners by the BOLD reduction can be
viewed as an online interval-partitioning problem. Each round $t$ corresponds to
an interval $[t,t+d_t+1)$, and overlapping intervals must be assigned to
different learners. This connects our scheduling-based approach to classical
interval partitioning and online interval coloring
\cite{kleinberg2006algorithm,faigle1995note,kierstead1981extremal}; see also
\cite{kolen2007interval} for a survey of interval scheduling. Unlike standard
formulations, the endpoints of our intervals are unknown when they arrive.
Moreover, our objective is not merely to minimize the number of learners, but to
control the regret.

\section{Refined BOLD reduction: guarantees and limitations}
\label{app:assignment}

In this section, we develop a refinement of the BOLD reduction 
of \cite{joulani2013}, which can be applied
in the arbitrary delays setting. We first show that 
this refined reduction can obtain the fast $\order(\qmax)$
rate when used with cooperating learners, while also obtaining 
$\order\brb{\sqrt{(T+\dtot)(1+\ln\qmax)}}$ regret against 
an arbitrary opponent. We then show that the sub-optimal $\sqrt{1+\ln\qmax}$
penalty can not be avoided by choosing any alternative scheduling, 
demonstrating that the black-box approach is insufficient for 
obtaining our desired best-of-both-worlds guarantees.

\subsection{A refined BOLD reduction}

\Cref{alg:refined-bold} implements the online partition rule of
\cref{sec:bold} using instances $\boldAlg_1,\ldots,\boldAlg_r$,
indexed in order of creation.
In the pseudocode, $r$ denotes the number of instances created so far.
Instance $\boldAlg_k$ handles the subsequence
$I_k=\{t\in[T]:i_t=k\}$, where $i_t$ is the index selected at round $t$.
Each instance maintains its own state and local round counter
and is idle when it has no pending feedback.

\begin{algorithm2e}[ht]
\caption{Refined BOLD}
\label{alg:refined-bold}
\KwInput{undelayed online learner $\boldAlg$.}
\KwInit{$r\gets0$.}
\For{$t=1,\dots,T$}{
  $i_t\gets1$\;
  \While{$i_t\le r$ and instance $\boldAlg_{i_t}$ is busy}{
    $i_t\gets i_t+1$\;
  }
  \If{$i_t=r+1$}{
    $r\gets r+1$\;
    initialize a fresh instance $\boldAlg_r$\;
  }
  obtain $w_t$ from $\boldAlg_{i_t}$, mark that instance busy, and play $w_t$\;
  observe $g_s$ for every $s\in\obs_t$ at the end of round $t$\;
  \ForEach{$s\in\obs_t$}{
    feed $g_s$ to $\boldAlg_{i_s}$, perform its local update, and mark it idle\;
  }
}
\end{algorithm2e}

\subsection{The adversarial guarantee}

\begin{proof}[Proof of \Cref{thm:assign}(i)]
  Let $i_t$ be the index selected at round $t$. For each $k$, let $I_k=\{t\in[T]:i_t=k\}$ be the subsequence handled by instance $\boldAlg_k$, and set $n_k=|I_k|$.
  Every busy instance holds exactly one pending round. Since there are $\sigma_t$ busy instances, the smallest-idle-index rule selects an index $i_t\le\sigma_t+1=q_t\le\qmax$, whether it reuses an existing instance or creates a new one.
  Consequently, we have
  \begin{equation}
    \sum_{k=1}^{\qmax} k n_k
    =
    \sum_{k=1}^{\qmax}k\sum_{t=1}^T\Ind\Set{i_t=k}
    =
    \sum_{t=1}^T\sum_{k=1}^{\qmax}k\Ind\Set{i_t=k}
    =
    \sum_{t=1}^T i_t
    \le
    \sum_{t=1}^T q_t=T+\dtot \;,
    \label{eq:bold-sum}
  \end{equation}
  where the last equality uses \Cref{fact:sets}(d).
  Each instance receives feedback before choosing its next action, so its local interaction is undelayed.
  Hence, applying the base learner's regret guarantee to each instance and applying 
  \Cref{eq:bold-sum} yields
  \begin{align*}
      \Reg_T
      &\le
      C_{\boldAlg}\sum_{k=1}^{\qmax}\sqrt{n_k}
      =
      C_{\boldAlg}\sum_{k=1}^{\qmax}\frac{\sqrt{k n_k}}{\sqrt k}
      \\ &\le
      C_{\boldAlg}\sqrt{\left(\sum_{k=1}^{\qmax}k n_k\right)
                          \left(\sum_{k=1}^{\qmax}\frac1k\right)}
      \le
      C_{\boldAlg}\sqrt{(T+\dtot)(1+\ln\qmax)} \;,
  \end{align*}
  where we have observed that $\sum_{k=1}^{\qmax}\frac1k \le 1+\ln\qmax$.
\end{proof}

\subsection{The game guarantee}

\begin{proof}[Proof of \Cref{thm:assign}(ii)]
Both players use the same deterministic smallest-idle-index rule.
Under common delays, their sets of busy instances coincide at the start of every round, so they select the same index.
Thus instance $k$ of one player interacts only with instance $k$ of the other.
Each instance receives feedback before choosing its next action and runs $\boldAlg$ using its local round counter, so each pair plays an undelayed copy of the original game.

Let $n_k$ be the number of rounds assigned to pair $k$, and denote its individual and social regrets by $\Reg_{n_k}^{p,(k)}$ and $\SocReg_{n_k}^{(k)}$.
By the assumed guarantee of $\boldAlg$ in part~(ii), these regrets are $\mathcal O(LM+G\sqrt M)$, uniformly over all pairs and local horizons $n_k$.
For any fixed comparator, regret decomposes across the local subsequences.
Taking the supremum for each player gives
\[
  \Reg_T^p \le \sum_k\Reg_{n_k}^{p,(k)} \;,
  \qquad
  \SocReg_T \le \sum_k\SocReg_{n_k}^{(k)} \;.
\]
There are at most $\qmax$ pairs, as shown in the proof of part~(i).
Summing their regret bounds therefore yields
\[
    \SocReg_T, \Reg_T^p
    =
    O\Brb{(LM+G\sqrt M)\qmax} \;. \qedhere
\]
\end{proof}

\subsection{A lower bound for every assignment}

\begin{proof}[Proof of the load bound in \Cref{prop:assign-lower}]
Fix an integer $\smax\ge1$ and write $q \defeq \qmax = \smax+1$.
Choose a positive integer $N$ divisible by $1^2,\ldots,q^2$, and define
\[
  \lambda_k=\frac{N}{k^2}\quad(k\in[q]),\qquad
  \lambda_{q+1}=0,\qquad
  b_j=\lambda_j-\lambda_{j+1}\quad(j\in[q]).
\]
For each $j\in[q]$, form $b_j$ blocks of $j$ consecutive rounds and concatenate all blocks in any fixed order.
Within a block of length $j$, release all feedback at the end of its last round, so its $a$th round has delay $j-a$.
No feedback remains pending between blocks.
Every valid assignment must use distinct instances for the rounds within a block, since no feedback from that block arrives before its final action is chosen.
Because $b_q=\lambda_q>0$, the largest block has length $q$, and the maximum number of pending rounds is exactly $q-1=\smax$.

The choice of $b_j$ ensures that,
for every $k\in[q]$, the number of blocks of length at least $k$ is
\begin{align*}
\sum_{j=k}^q b_j
&=\sum_{j=k}^q(\lambda_j-\lambda_{j+1})
  =\lambda_k.
\end{align*}
Moreover, since there are $b_{j}$ blocks of length $j$ by construction, we have
\begin{align*}
    T = \sum_{j=1}^{q}j b_{j}.
\end{align*}
Hence, the total number of rounds can be counted in two equivalent ways: by summing the block lengths, or by counting the blocks containing a $k$-th round for each $k\in[q]$:
\begin{align}
    T=\sum_{j=1}^q jb_j
    =\sum_{k=1}^q\sum_{j=k}^q b_j
    =\sum_{k=1}^q\lambda_k \;.
    \label{eq:assignment-loads-1}
\end{align}
A block of length $j$ contributes $j$ to $T$.
Its $a$th round has delay $j-a$, so its contribution to
$\dtot$ is $\sum_{a=1}^j(j-a)=j(j-1)/2$.
Hence, its contribution to $T+\dtot$ is
\[
    j+\frac{j(j-1)}2
    =\frac{j(j+1)}2
    =\sum_{k=1}^j k.
\]
Summing over all blocks and exchanging the order of summation,
we obtain
\begin{align}
    T+\dtot
    =\sum_{j=1}^q b_j\sum_{k=1}^j k
      =\sum_{k=1}^q k\sum_{j=k}^q b_j 
    =\sum_{k=1}^q k\lambda_k
      =N\sum_{k=1}^q\frac1k
      =NH_q,
    \label{eq:assignment-loads-2}
\end{align}
where $H_q=\sum_{k=1}^q 1/k$.
Now consider any valid assignment, allowing any number of instances, and order its loads as $n_1\ge n_2\ge\cdots$. Since no feedback arrives before the end of a block, each instance can be assigned at most one round within that block. Hence, any $r$ instances can cover at most $\min\{r,j\}$ rounds in a block of length $j$. Summing over blocks and again using \Cref{eq:assignment-loads-1}, we obtain
\begin{align*}
  \sum_{i=1}^r n_i
  \le \sum_{j=1}^q b_j\min\{r,j\}
  =\sum_{k=1}^{\min\{r,q\}}\sum_{j=k}^q b_j
  =\sum_{k=1}^{\min\{r,q\}}\lambda_k.
\end{align*}
Let $m\ge q$ be the number of instances assigned at least one round, where
$m\ge q$ follows from the existence of a block of length $q$. Extend the
sequence $(\lambda_k)_{k=1}^q$ by setting $\lambda_k=0$ for $q<k\le m$.
Since
both sequences are nonincreasing, have the same total sum (i.e., $\sum_{i=1}^m\lambda_{i}=\sum_{i=1}^mn_{i}=T$), and satisfy
\begin{align*}
  \sum_{i=1}^r n_i\leq\sum_{k=1}^r\lambda_k
  \qquad\forall r\in[m],
\end{align*}
we have that the vector $(\lambda_{1},\ldots,\lambda_{m})$ majorizes $(n_{1},\ldots,n_{m})$, so
by concavity of square root and Karamata's inequality \citep[Proposition 3.C.1]{marshall1979inequalities}:
\begin{align*}
  \sum_{i=1}^m\sqrt{n_i}
  \geq\sum_{k=1}^m\sqrt{\lambda_k}
  =\sum_{k=1}^q\sqrt{\lambda_k}.
\end{align*}

Finally, substituting $\lambda_k=N/k^2$ and recalling that $T+\dtot=NH_q$ from \Cref{eq:assignment-loads-2}, we obtain
\begin{align}
    \sum_{i=1}^m\sqrt{n_i}
    &\ge \sqrt{N}H_q
    =\sqrt{(T+\dtot)H_q}
    \ge\sqrt{(T+\dtot)\ln(q+1)}
    \ge\sqrt{(T+\dtot)\frac{1+\ln\qmax}{2}} \;,
    \label{eq:assignment-sharp-lower}
\end{align}
where the third inequality uses $H_q \ge \ln(q+1)$, and the last follows from $q = \qmax \ge 1$.
\end{proof}

\subsection{Realizing the lower bound with optimistic OMD}
\label{app:assignment-omd-lower}

We consider the following adaptive single-anchor optimistic OMD learner, using the previous local gradient as its hint.
Within each instance, let $x_j$ and $g_j$ denote the action and gradient at local round $j$.
Set $g_0=0$, fix $\gamma>0$, and define
\[
    V_j = \sum_{i=1}^j\dnorm{g_i-g_{i-1}}^2 \;,
    \qquad
    c_j = \gamma+\sqrt{V_j/M} \;.
\]
After receiving $g_j$, the instance computes
\begin{equation}
\label{eq:assignment-optimistic-lower-update}
  x_{j+1}\in\argmin_{x\in\cW}
  \left\{\langle 2g_j-g_{j-1},x\rangle+c_jD_\psi(x\mid x_j)\right\}.
\end{equation}
Each instance maintains its own iterate, local counter, and accumulated variation $V_j$, without knowing its eventual number of rounds.
For the lower bound, take $\cW=[-1,1]$, $\psi(x)=x^2/2$, and $M=2$, and initialize every instance at $x_1=1$.

\begin{proof}[Proof of the optimistic OMD claim in \Cref{prop:assign-lower}]
  Use the block delays constructed above.
  After observing the played action $w_t$, the adversary chooses the linear loss $f_t(w)=g_t w$, where $g_t=1$ if $x_t\ge0$ and $g_t=-1$ otherwise.
  This adaptive choice is permitted by the online protocol of \Cref{sec:prelim}.
  It depends only on the played action, so the argument applies to every realized assignment, including those generated by a randomized rule.
  Against the common comparator $u=0$, the loss difference at round $t$ is $g_t x_t=|x_t|$.

  Consider an instance with $n\ge1$ local rounds.
  In its local notation, $g_jx_j=|x_j|$, $|g_j|=1$, and $V_j\le4j$.
  The update \eqref{eq:assignment-optimistic-lower-update} therefore becomes
  \[
    x_{j+1}=\operatorname{\Pi}_{[-1,1]}
    \left(x_j-\frac{2g_j-g_{j-1}}{c_j}\right),
    \qquad c_j\le\gamma+\sqrt{2j} \;,
  \]
  where $\operatorname{\Pi}_{[-1,1]}(x)=\argmin_{y\in [-1,1]} |y-x|^2$ denotes the Euclidean projection 
  onto $[-1,1]$.
  The quantity $2g_j-g_{j-1}$ has sign $g_j$ and magnitude at least one, including at $j=1$ because $g_0=0$.
  Thus the unprojected step moves toward zero and possibly beyond it.
  The distance to the boundary in this direction is $1+|x_j|$, so
  \[
    |x_{j+1}-x_j|
    =\min\left\{\frac{|2g_j-g_{j-1}|}{c_j},\,1+|x_j|\right\}
    \ge\min\{c_j^{-1},1\}
    \ge\frac{1}{(\gamma+\sqrt2)\sqrt j} \;.
  \]
  For $n\ge2$, the triangle inequality yields
  \begin{align*}
    \sum_{j=1}^{n-1}\abs{x_{j+1}-x_j} 
    \le \sum_{j=1}^{n-1} \abs{x_{j+1}}+\abs{x_j}
    =
      \abs{x_1} +2\sum_{j=2}^{n-1}\abs{x_j}  + \abs{x_n}
    \le 2\sum_{j=1}^n \abs{x_j},
  \end{align*}
  so applying this with the previous display yields
  \begin{align*}
    2\sum_{j=1}^n|x_j|
    &\ge\sum_{j=1}^{n-1}|x_{j+1}-x_j|
    \ge\frac{1}{\gamma+\sqrt2}\sum_{j=1}^{n-1}\frac1{\sqrt j}
    \ge\frac{n-1}{(\gamma+\sqrt2)\sqrt n}
    \ge\frac{\sqrt n}{2(\gamma+\sqrt2)}.
  \end{align*}
  Hence $\sum_{j=1}^n|x_j|\ge\sqrt n/(4(\gamma+\sqrt2))$.
  This bound also holds for $n=1$, since $x_1=1$.
  Summing over instances against the same comparator and applying \eqref{eq:assignment-sharp-lower} gives
  \[
    \Reg_T\ge\Reg_T(0)=\sum_{t=1}^T|x_t|
    \ge\frac{1}{4(\gamma+\sqrt2)}\sum_k\sqrt{n_k}
    \ge\frac{\sqrt{(T+\dtot)(1+\ln\qmax)}}{4\sqrt{2}(\gamma+\sqrt2)} \;.
  \]
  This inequality holds for every realization of the assignment rule.
  For each fixed $\gamma>0$, it implies the stated regret lower bound, with a constant independent of $T$ and $\qmax=\smax+1$.
\end{proof}

\section{Online convex optimization with arbitrary losses}
\label{app:oco}

In this section, we consider single-player online convex optimization with arbitrary convex losses and arbitrary delays, and we show that \MAO with optimism weight $\gamma=2G/\sqrt M$ achieves the optimal regret rate under delays.

\paragraph{Setting.}
At each round $t\in[T]$, the learner chooses $w_t\in\cW$, and an adversary chooses a convex loss $f_t:\cW\to\R$.
The learner incurs the loss $f_t(w_t)$, and a subgradient $g_t\in\partial f_t(w_t)$ is revealed at the end of round $t+d_t$, where the delays $d_t$ are arbitrary and unknown.
We assume that $\dnorm{g_t}\le G$ for every $t\in[T]$ and that the learner knows $G$, and the regret is $\Reg_T\defeq\sup_{u\in\cW}\sumtT\brb{f_t(w_t)-f_t(u)}$.
We impose no smoothness assumptions on the loss sequence.

\paragraph{Algorithm.}
The learner runs \Cref{alg:mao} with a regularizer $\psi$ satisfying \Cref{ass:diam} and optimism weight $\gamma=2G/\sqrt M$.
With this choice, the first term in \Cref{eq:master-gamma}
is $2\gamma M\qmax=4G\sqrt M\,\qmax$.

\begin{corollary}
\label{cor:oco}
     For arbitrary convex losses with $\dnorm{g_t}\le G$ and arbitrary unknown delays, \textnormal{\MAO} with optimism weight $\gamma=2G/\sqrt M$ guarantees $\Reg_T=\order\brb{G\sqrt{M(T+\dtot)}}$.
\end{corollary}

\begin{proof}
     Each hint is an observed subgradient or zero, so
     $\dnorm{g_t-h_t}\le2G$.
     By \Cref{fact:kappa}(b),
     $\sumtT\kappa_t\dnorm{g_t-h_t}^2\le4G^2(T+2\dtot)$.
     By convexity, regret is bounded by linearized regret.
     Applying \eqref{eq:master-gamma} with
     $\gamma=2G/\sqrt M$ and dropping the nonpositive
     stability term yields
     \begin{align*}
     \Reg_T
     &\le 4G\sqrt M\,\qmax
          +4\sqrt2\,G\sqrt{M(T+2\dtot)}\\
     &\le (4+4\sqrt2)G\sqrt{M(T+2\dtot)}\\
     &\le (8+4\sqrt2)G\sqrt{M(T+\dtot)},
     \end{align*}
     where the second inequality uses
     $\qmax\le\sqrt{T+2\dtot}$ from \Cref{fact:sets}(e),
     and the last inequality uses
     $T+2\dtot\le2(T+\dtot)$.
\end{proof}

The dependence on $T+\dtot$ matches the optimal regret rate of online learning with delays~\cite{quanrud2015}.

\section{Auxiliary results}
\label{app:auxiliary}\label{app:delay}

We collect basic properties of the delay sets and congestion quantities defined in \cref{sec:prelim}.

\begin{fact}[Delay-set identities] \label{fact:sets}
    The following properties hold for every $t\in[T]$:
    \begin{enumerate}[(a),itemsep=0pt]
        \item $\hist_t\sqcup\miss_t=[t-1]$, $\hist_{t+1}=\hist_t\sqcup\obs_t$, $\act_t=\miss_{t+1}\sqcup\obs_t$;
        \item $\obs_1,\dots,\obs_T$ partition $[T]$;
        \item $q_{t+1}=q_t+1-\abs{\obs_t}\le q_t+1$, $q_1=1$, $\miss_1=\miss_{T+1}=\emptyset$;
        \item $\sum_{t\in[T]}\sigma_t=\dtot$ under the assumption that $t+d_t\le T$ for every $t$;
        \item $\qmax^2\le T+2\dtot$.
    \end{enumerate}
\end{fact}

\begin{proof}[Proof of \cref{fact:sets}]
(a) A round $s<t$ has either $s+d_s<t$ or $s+d_s\ge t$; the elements of $\hist_{t+1}\setminus\hist_t$ are the $s\le t$ with $s+d_s=t$; and $\act_t=\{s\le t:s+d_s\ge t\}$ splits according to $s+d_s=t$ (then $s\in\obs_t$) or $s+d_s\ge t+1$ (then $s\in\miss_{t+1}$).
(b) Each $s$ lies in $\obs_{s+d_s}$ and in no other $\obs_t$.
(c) By (a), $q_t=\abs{\miss_{t+1}}+\abs{\obs_t}=q_{t+1}-1+\abs{\obs_t}$; $\miss_1=\emptyset$ because no round precedes round $1$, and $\miss_{T+1}=\emptyset$ because every gradient arrives by the end of round $T$.
(d) $\sum_t\sigma_t=\sum_t\sum_s\Ind\{s<t\le s+d_s\}=\sum_sd_s$.
(e) Let $\sigma_t=\smax=k$. Since $\sigma_1=0$ and $\sigma$ grows by at most one per round by (c), $t\ge k+1$ and $\sigma_{t-j}\ge k-j$ for $0\le j\le k$, so $\dtot\ge\sum_{j=0}^{k}(k-j)=k(k+1)/2$ by (d), and $T\ge k+1$. Hence $\qmax^2=k(k+1)+(k+1)\le2\dtot+T$.
\end{proof}

\begin{fact}[Congestion properties]
\label{fact:kappa}
The quantities $\kappa_{s,t}$ and $\kappa_s$ satisfy the following properties:
\begin{enumerate}[(a),itemsep=0pt]
    \item We have $\kappa_{t,t}=q_t$, and $t\mapsto\kappa_{s,t}$ is nondecreasing for each fixed $s$. Moreover,
    \begin{equation*}
        q_t\le\kappa_{s,t}\le\kappa_s\le\qmax
        \qquad\text{for every }s\in\act_t,
    \end{equation*}
    with $\kappa_{s,t}=\kappa_s$ whenever $s\in\obs_t$.
    \item The following bound holds:
    \begin{equation*}
        \sumtT\kappa_t\le T+2\dtot \;.
    \end{equation*}
    Consequently, for all $a_1,\dots,a_T\ge0$,
    \begin{equation*}
        \sumtT \kappa_t a_t
        \le
        \min\bbcb{
            \qmax\sumtT a_t,\;
            (T+2\dtot)\max_{t\in[T]}a_t
        } \;.
    \end{equation*}
\end{enumerate}
\end{fact}

\begin{proof}[Proof of \cref{fact:kappa}]
(a) is immediate from the definition.
(b) A round pending at some $r\in\{s,\dots,s+d_s\}$ was either pending at round $s$ or is one of the rounds $s,\dots,r-1$, so $q_r\le1+\sigma_s+(r-s)\le1+\sigma_s+d_s$ and $\kappa_s\le1+\sigma_s+d_s$; summing and using \cref{fact:sets}(d) gives $\sum_s\kappa_s\le T+2\dtot$. The second claim follows from $\kappa_s\le\qmax$ and $a_s\le\max_ra_r$.
\end{proof}

\newcommand{\cl}[2]{\lambda_{#1}^{(#2)}}
\begin{lemma}\label{fact:variance-sum-swap}\label{lem:variance-sum-swap}
    Let $(\delta_t)_t$ and $(\cl{s}{t})_{s,t}$ be arbitrary non-negative sequences of real numbers. Then
    \begin{align*}
        \sumtT \sum_{s\in\obs_t}\cl{s}{\tpp}\delta_s
        =
        \sumtT \cl{t}{t+d_t+1}\delta_t
    \end{align*}
\end{lemma}
\begin{proof}
    Observe that by definition of 
    $\obs_t=\Set{s\in[t]: s+d_s=t}$, we have
    \begin{align*}
        \sumtT \sum_{s\in\obs_t}\cl{s}{\tpp}\delta_s
        &=
        \sumtT \sum_{s=1}^t\boldsymbol{1}\Set{t=s+d_s}\cl{s}{\tpp}\delta_s\\
        &=
        \sum_{s=1}^T \sum_{t=s}^T\boldsymbol{1}\Set{t=s+d_s}\cl{s}{\tpp}\delta_s\\
        &=
        \sumtT\cl{t}{t+d_t+1}\delta_t
    \end{align*}
    where the last line observes that the indicator is non-zero only at $t=s+d_s$.
\end{proof}

\begin{lemma} \label{lem:learning-rate-sum}
    Let $\lambda_1, \dots, \lambda_T \ge 0$, let $\gamma \ge \sqrt{\max_t \lambda_t} > 0$, and let $\alpha_1, \dots, \alpha_T > 0$ satisfy $\alpha_t \le 1/q_t$ for every $t \in [T]$.
    Define $0 = \eta_1 \le \dots \le \eta_{T+1}$ such that, for all $t \in [T]$,
    \begin{equation*}
        \eta_{t+1} = \eta_t + \sum_{s \in \obs_t} \frac{\lambda_s}{\gamma + \eta_s\alpha_s} = \sum_{s \in \hist_{t+1}} \frac{\lambda_s}{\gamma + \eta_s\alpha_s} \;.
    \end{equation*}
    Then,
    \begin{equation*}
        \eta_{T+1}
        =
        \sumtT \frac{\lambda_t}{\gamma + \eta_t\alpha_t}
        \le
        \sqrt{2\sumtT \frac{\lambda_t}{\alpha_t}} \;.
    \end{equation*}
\end{lemma}
\begin{proof}
    For every $t \in [T]$, define $z_t \defeq \frac{\lambda_t}{\gamma + \eta_t\alpha_t}$ and observe that $\eta_t = \sum_{s\in\hist_t}z_s$ by the fact that $\hist_t = \obs_1 \sqcup \dots \sqcup \obs_{t-1}$.
    By assumption, we have that $z_t \le \frac{\lambda_t}{\gamma} \le \gamma$.
    Now let $Z_t \defeq \sum_{k=1}^t z_k$ for all $t \in \{0, \dots, T\}$ and observe that, when $t \ge 1$,
    \begin{equation*}
        Z_{t-1}
        =
        \eta_t + \sum_{s<t: s\notin\hist_t} z_s
        =
        \eta_t + \sum_{s \in \miss_t} z_s
        \le
        \eta_t + \gamma\sigma_t \;.
    \end{equation*}
    Therefore, by telescoping,
    \begin{align*}
        \eta_{T+1}^2
        &=
        \sumtT \brb{Z_t^2 - Z_{t-1}^2}
        \\ &=
        \sumtT (Z_t - Z_{t-1})(Z_t + Z_{t-1})
        \\ &=
        \sumtT z_t(2Z_{t-1} + z_t)
        && \text{($Z_t = Z_{t-1} + z_t$)}
        \\ &\le
        2\sumtT z_t(\eta_t + \gamma\sigma_t + \gamma)
        && \text{($Z_{t-1} \le \eta_t+\gamma\sigma_t$ and $z_t \le \gamma$)}
        \\ &=
        2\sumtT z_t(\eta_t + \gamma q_t)
        && \text{($q_t = \sigma_t+1$)}
        \\ &\le
        2\sumtT z_t\bbrb{\eta_t + \frac{\gamma}{\alpha_t}}
        && \text{($\alpha_t \le 1/q_t$)}
        \\ &=
        2\sumtT \frac{\lambda_t}{\alpha_t} \;.
    \end{align*}
    Taking the square root on both sides proves the claim.
\end{proof}

\end{document}